\documentclass{article}
\usepackage{spconf,amsmath,graphicx}
\usepackage{hyperref}
\usepackage{xspace}
\usepackage[export]{adjustbox}
\usepackage{booktabs, multirow}	
\usepackage{color, colortbl}
\usepackage{cleveref}
\newcommand{\etal}{\textit{et al}.}

\hypersetup{hidelinks}

\usepackage[disable]{todonotes}

\newcommand{\fnum}[1]{\emph{f/#1}}
\newcommand{\fstop}{\emph{f}-stop\xspace}
\newcommand{\fstops}{\emph{f}-stops\xspace}
\newcommand{\dpdd}{DPDD$_S$}
\newcommand{\task}{Single-Image Defocus Deblurring}
\newcommand{\stask}{SIDD}

\definecolor{lgray}{gray}{0.925}
\definecolor{purp}{rgb}{0.75, 0.20, 0.55}

\usepackage{indentfirst}

\title{The RealDefocus Benchmark for Defocus Deblurring}
\name{Tim Seizinger, \quad
Zhuyun Zhou$^*$\thanks{*Corresponding author}, \quad
Radu Timofte\thanks{This work was supported by the Alexander von Humboldt Foundation.}
}
\address{Computer Vision Lab, CAIDAS \& IFI, University of W\"urzburg, Germany}

\begin{document}

\maketitle

\begin{abstract}

Single-Image Defocus Deblurring (SIDD) aims to recover an all-in-focus image from a single defocused observation, but rigorous and reproducible evaluation remains challenging due to the scarcity of realistic, high-resolution datasets with well-aligned defocused/sharp pairs and standardized protocols.
We build on RealDefocus, a benchmark derived from the real-world RealBokeh dataset originally proposed for Bokeh Rendering. RealDefocus provides paired defocused inputs and sharp ground truth images, predefined training/validation/test splits, and a unified evaluation framework for comparing image restoration and neural rendering approaches. We further outline a benchmarking protocol with cross-dataset validation to assess reconstruction quality and generalization.
The project page is publicly available at: \url{www.github.com/TimSeizinger/RealDefocus-Benchmark}.
\end{abstract}


\begin{figure*}[t]
    \centering
    \setlength{\tabcolsep}{1pt}
    \small
    \renewcommand{\arraystretch}{0.65}
    \def\widthcomp{0.19}
    \begin{tabular}{cccccc}
    Sharp GT \fnum{22.0} & Blurry Input \fnum{8.0} & Blurry Input \fnum{4.0} & Blurry Input \fnum{3.2} & Blurry Input \fnum{2.0}  \\
    \addlinespace[0.75pt]
    \rotatebox{90}{\hspace{2mm}Lens: 61mm}
    \includegraphics[width=\widthcomp\linewidth, trim={0 100px 0 200px},clip]{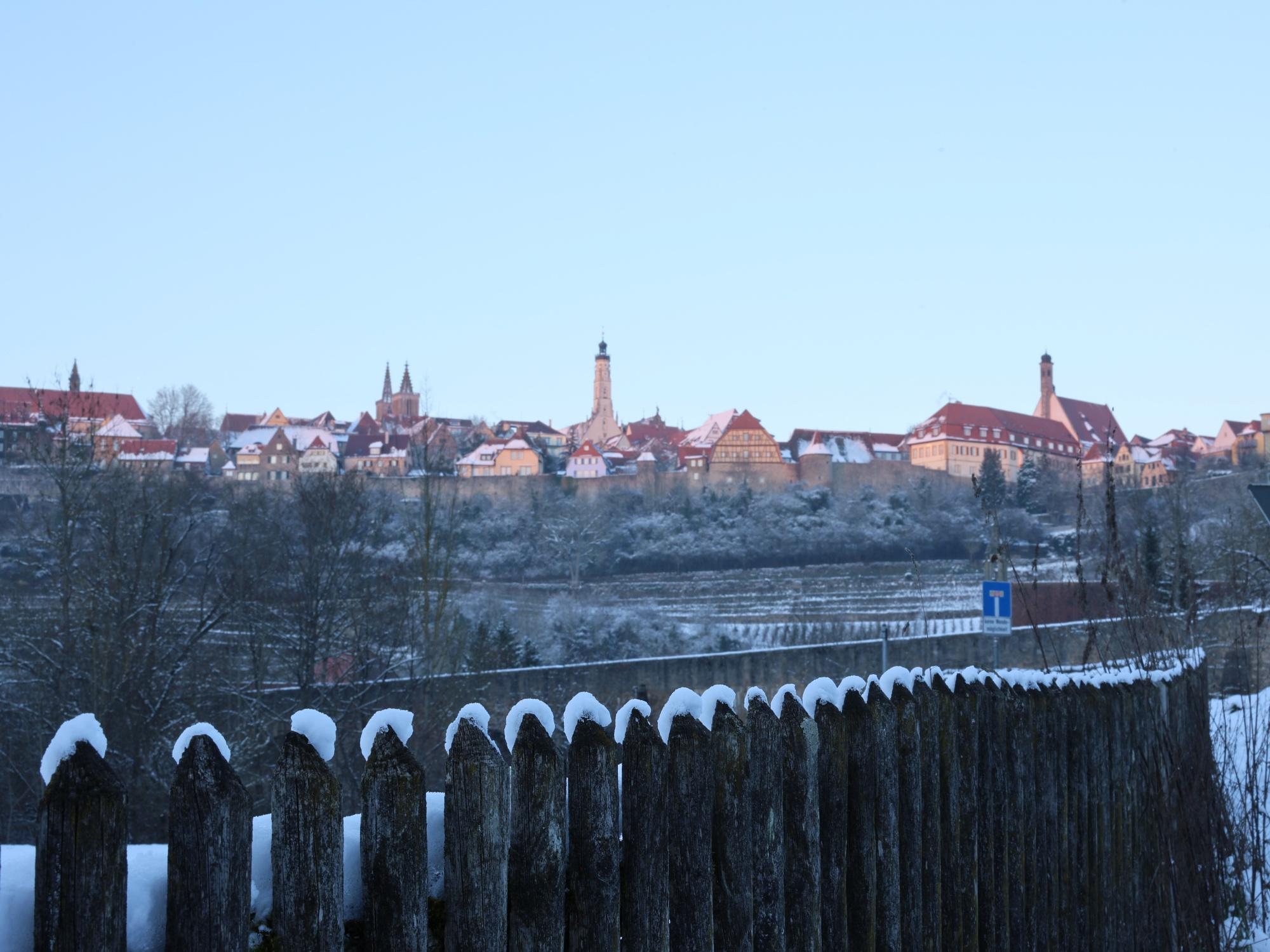} &     
    \includegraphics[width=\widthcomp\linewidth, trim={0 100px 0 200px},clip]{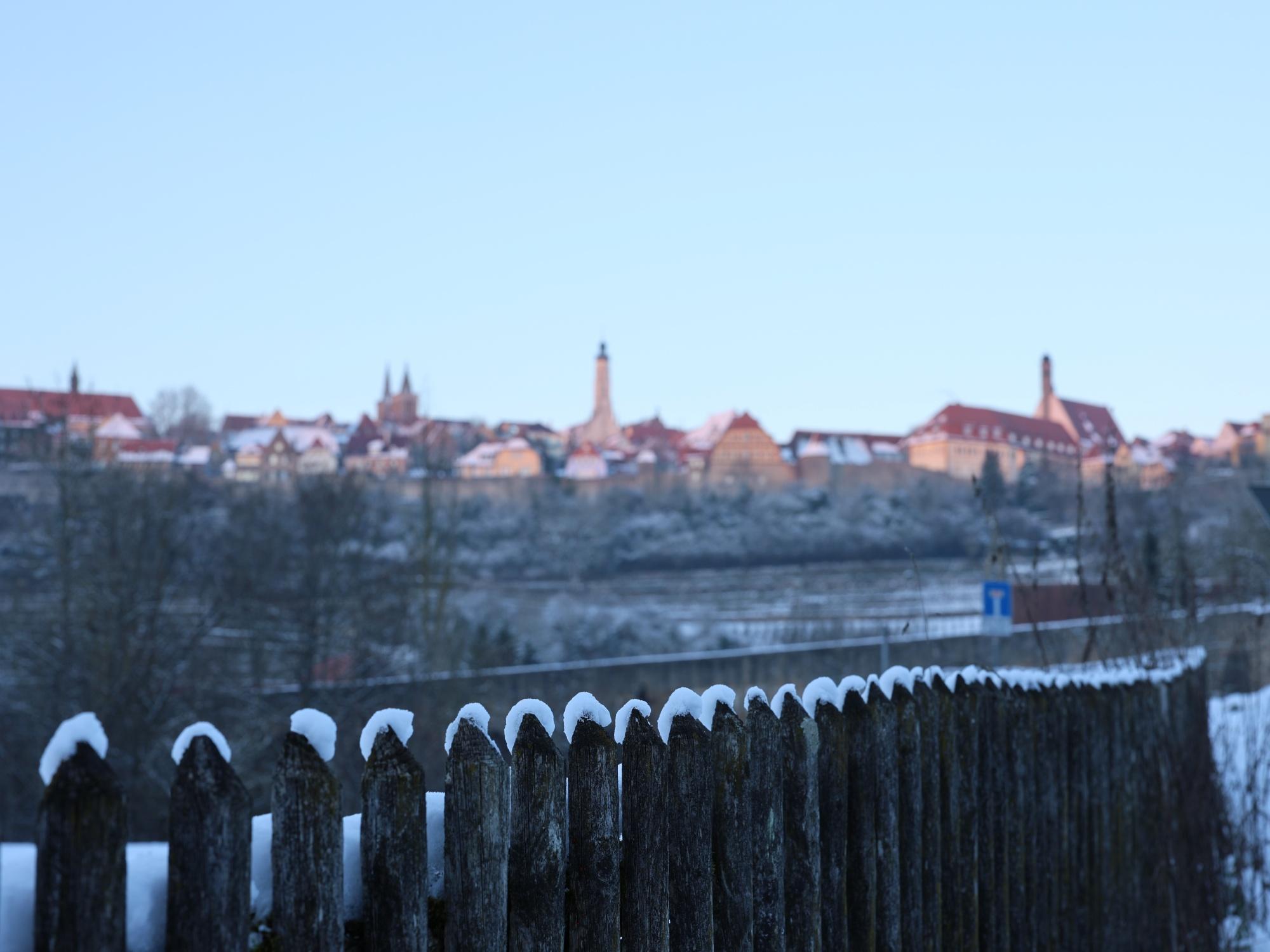} & 
    \includegraphics[width=\widthcomp\linewidth, trim={0 100px 0 200px},clip]{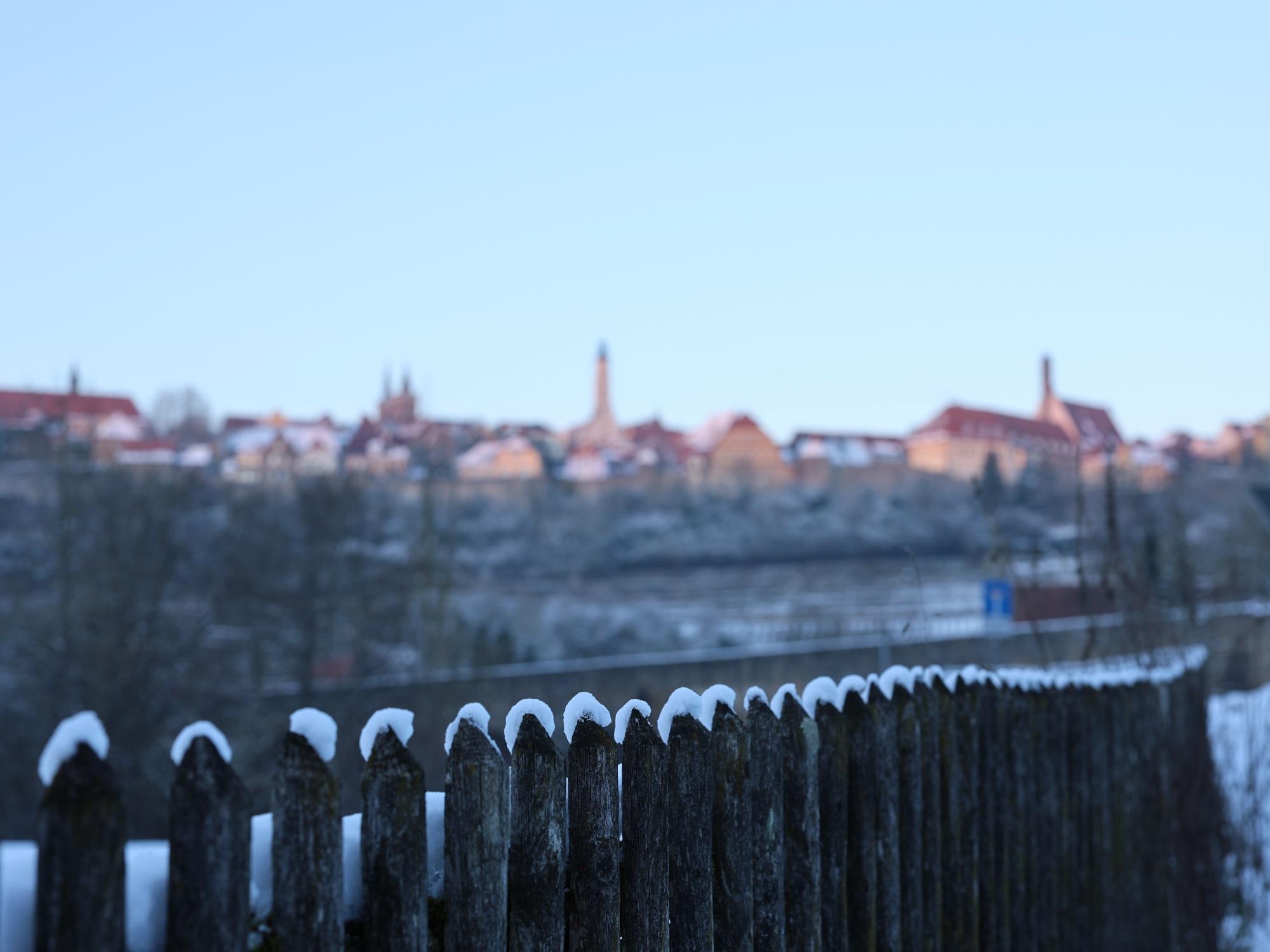} &  
    \includegraphics[width=\widthcomp\linewidth, trim={0 100px 0 200px},clip]{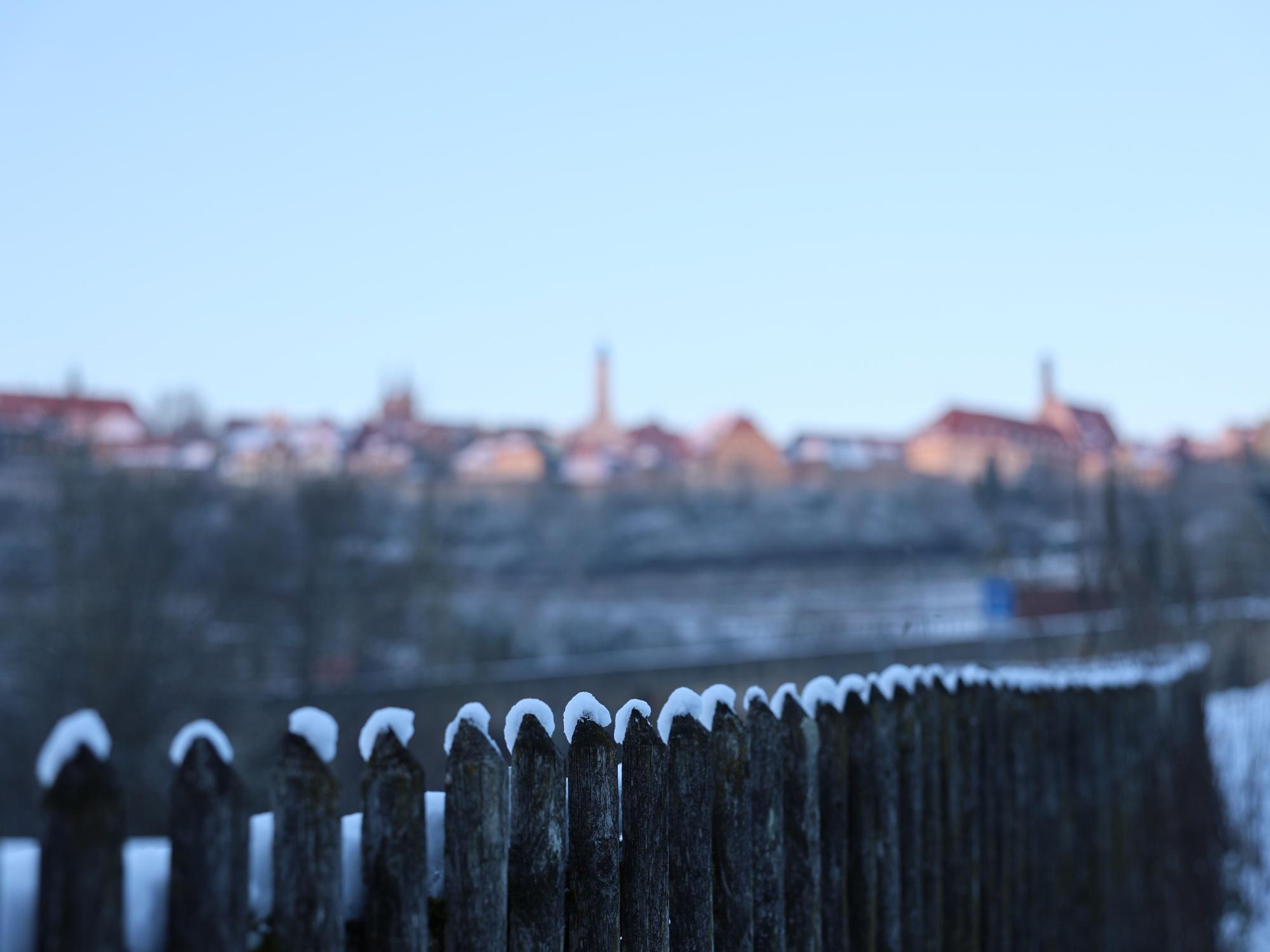} &
    \includegraphics[width=\widthcomp\linewidth, trim={0 100px 0 200px},clip]{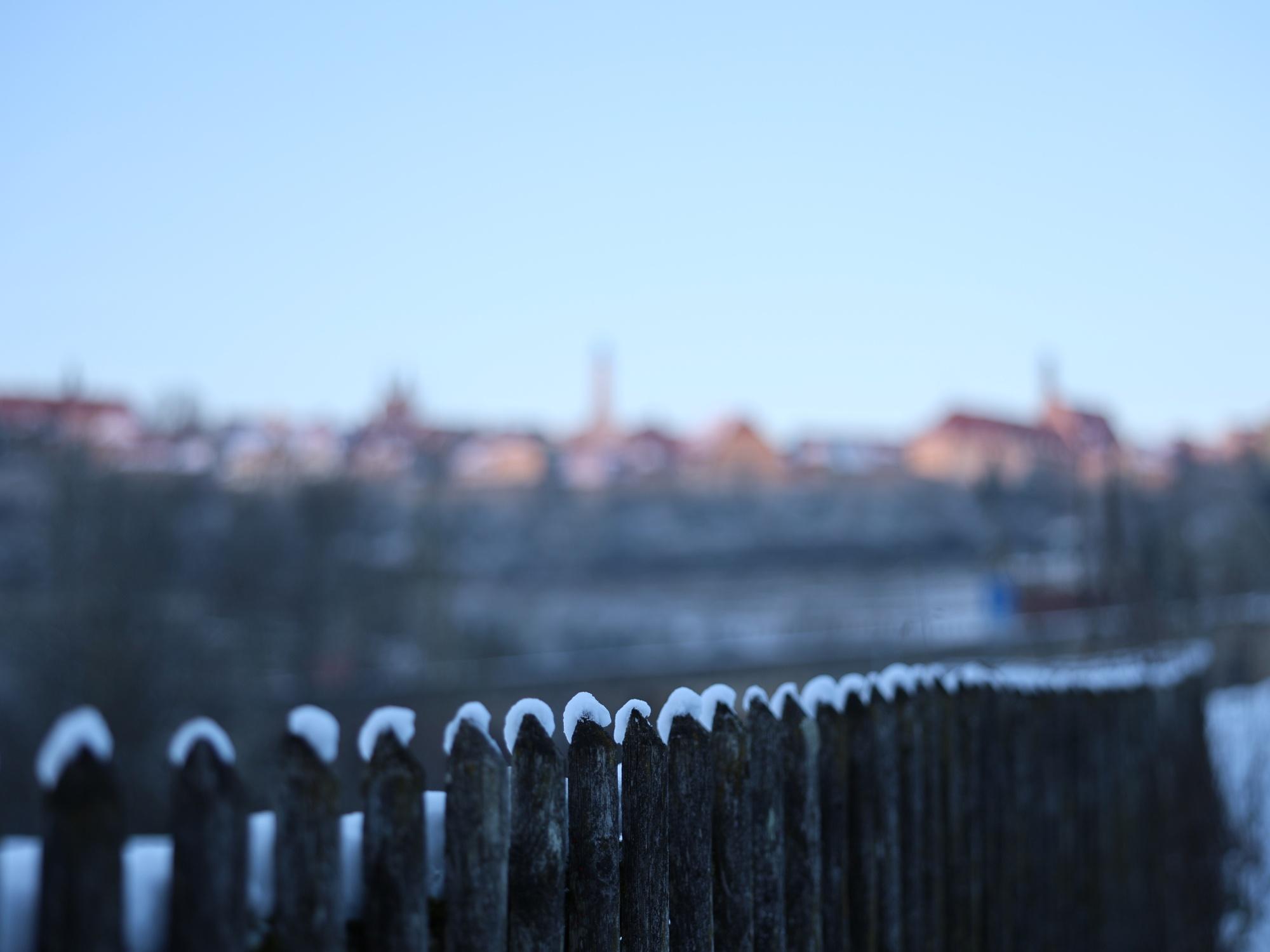} \\
    
    \addlinespace[0.75pt]
    \rotatebox{90}{\hspace{2mm}Lens: 70mm}
    \includegraphics[width=\widthcomp\linewidth, trim={0 100px 0 200px},clip]{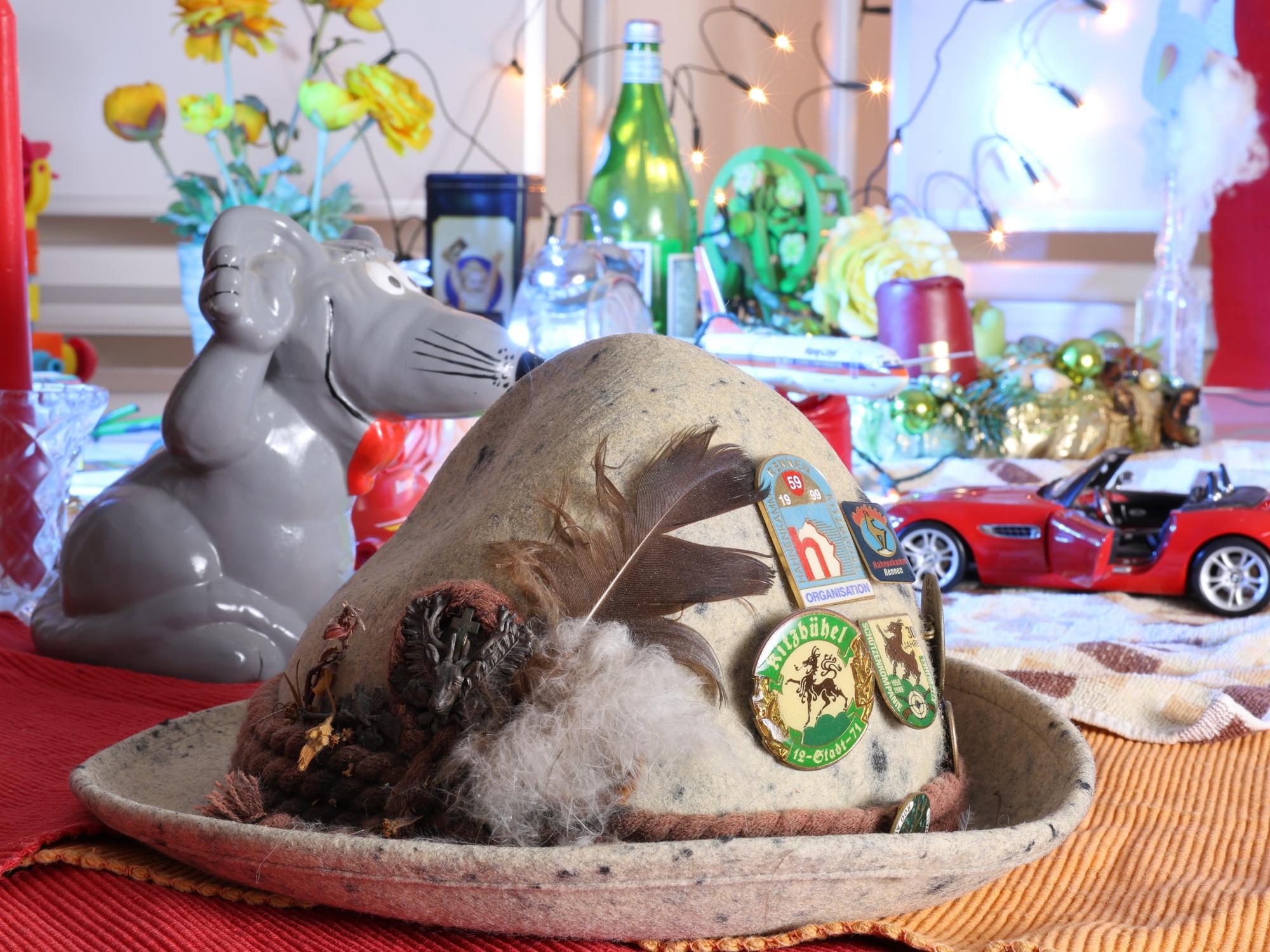} &     
    \includegraphics[width=\widthcomp\linewidth, trim={0 100px 0 200px},clip]{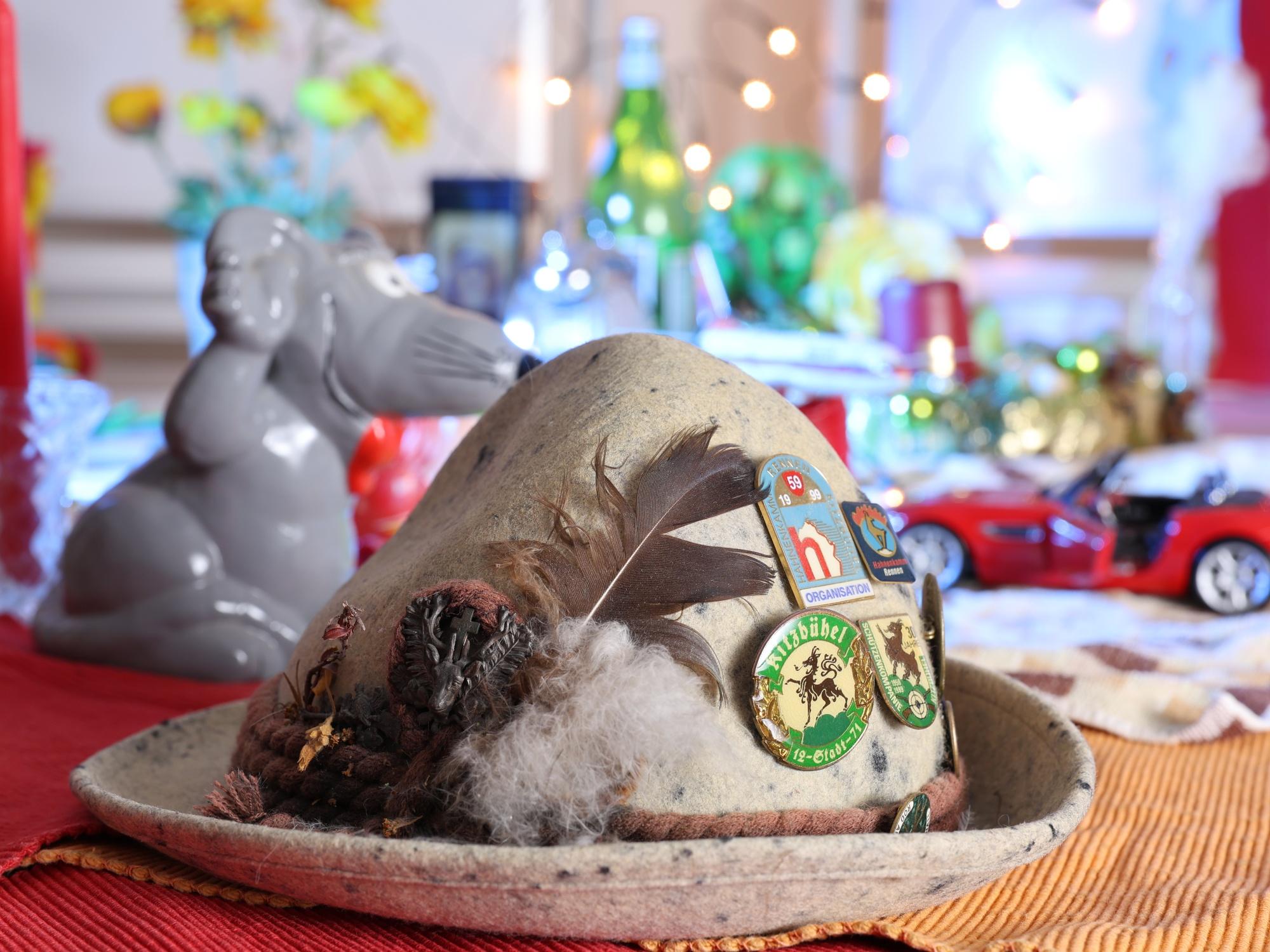} & 
    \includegraphics[width=\widthcomp\linewidth, trim={0 100px 0 200px},clip]{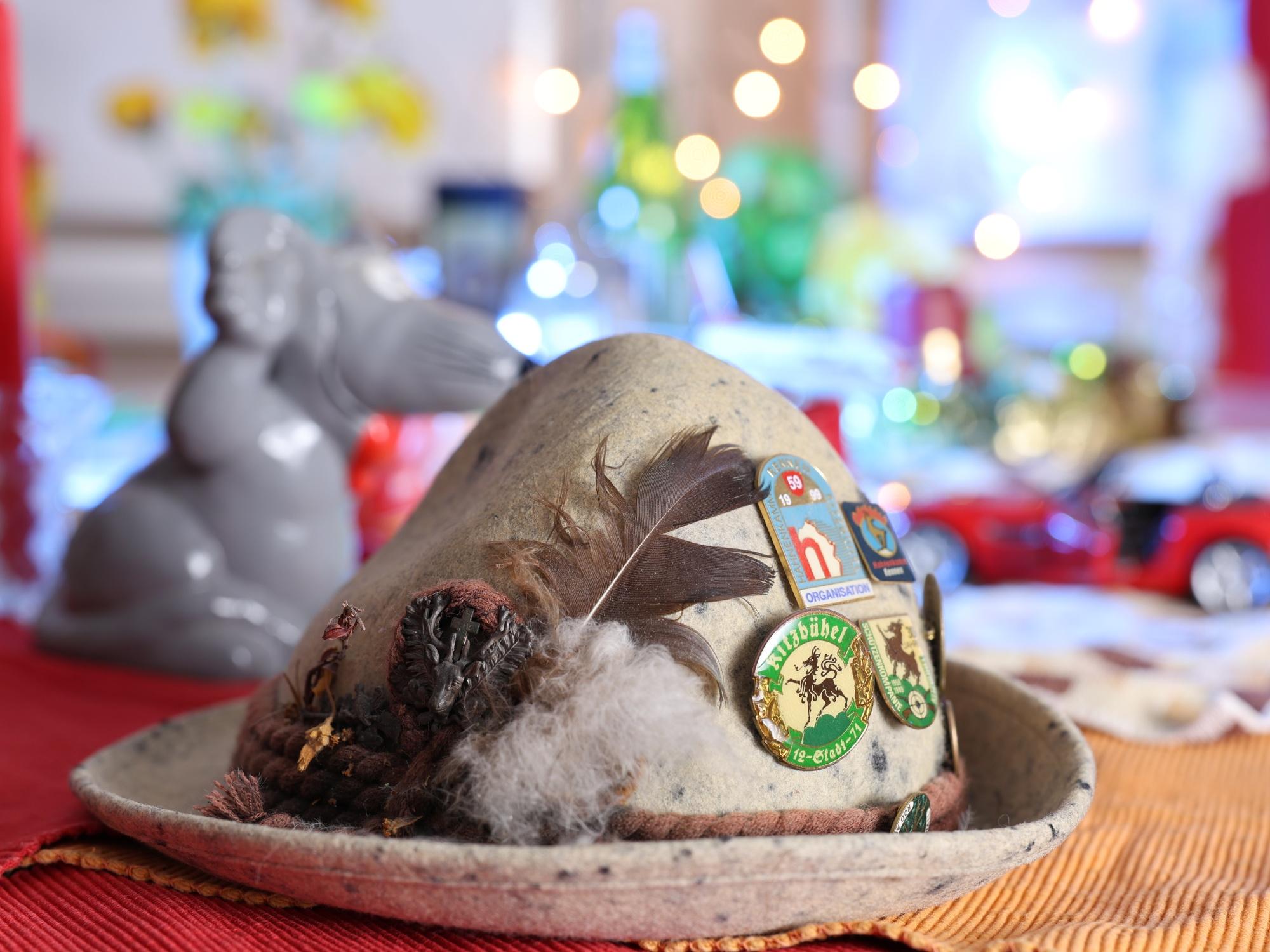} &  
    \includegraphics[width=\widthcomp\linewidth, trim={0 100px 0 200px},clip]{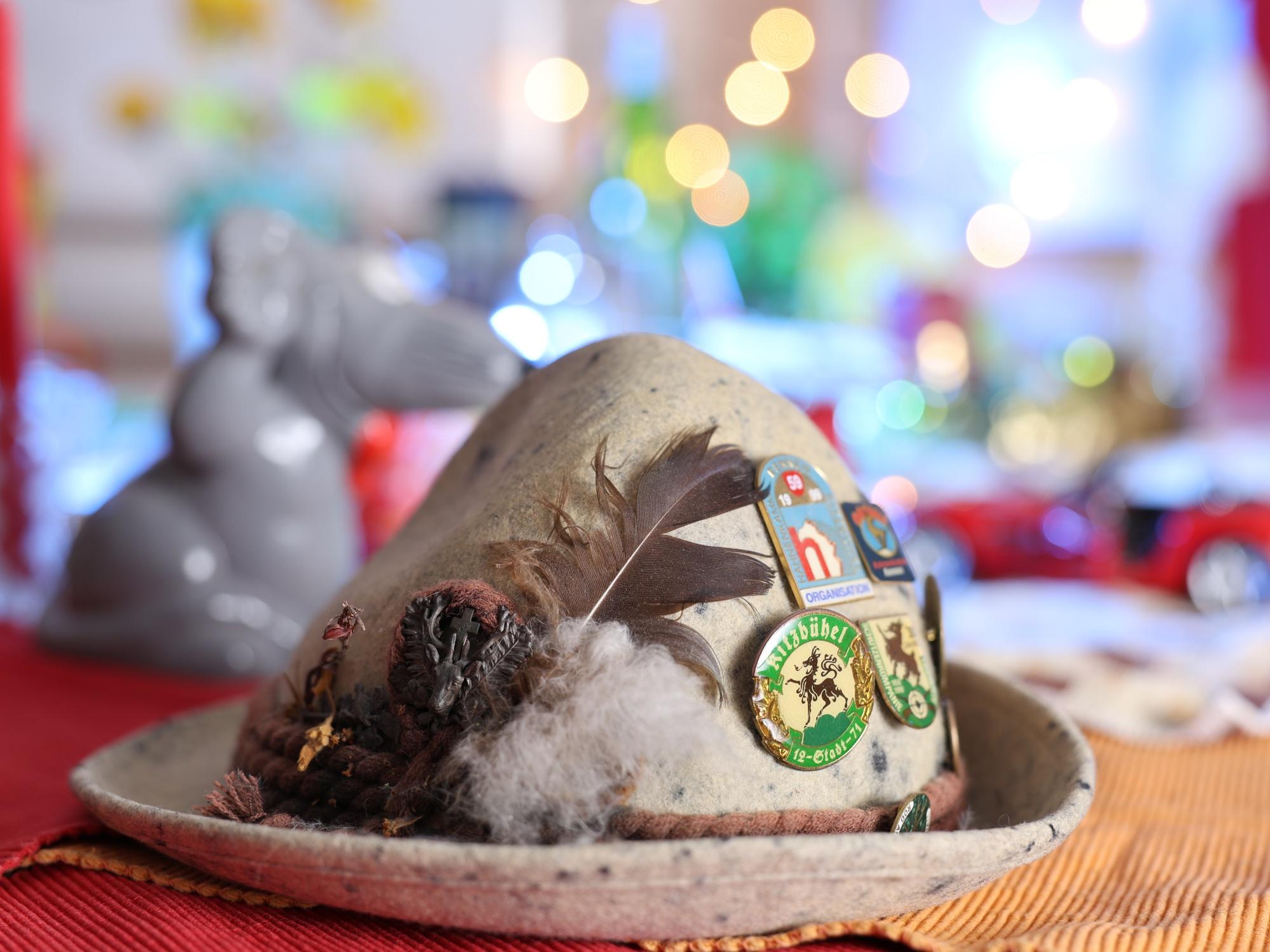} &
    \includegraphics[width=\widthcomp\linewidth, trim={0 100px 0 200px},clip]{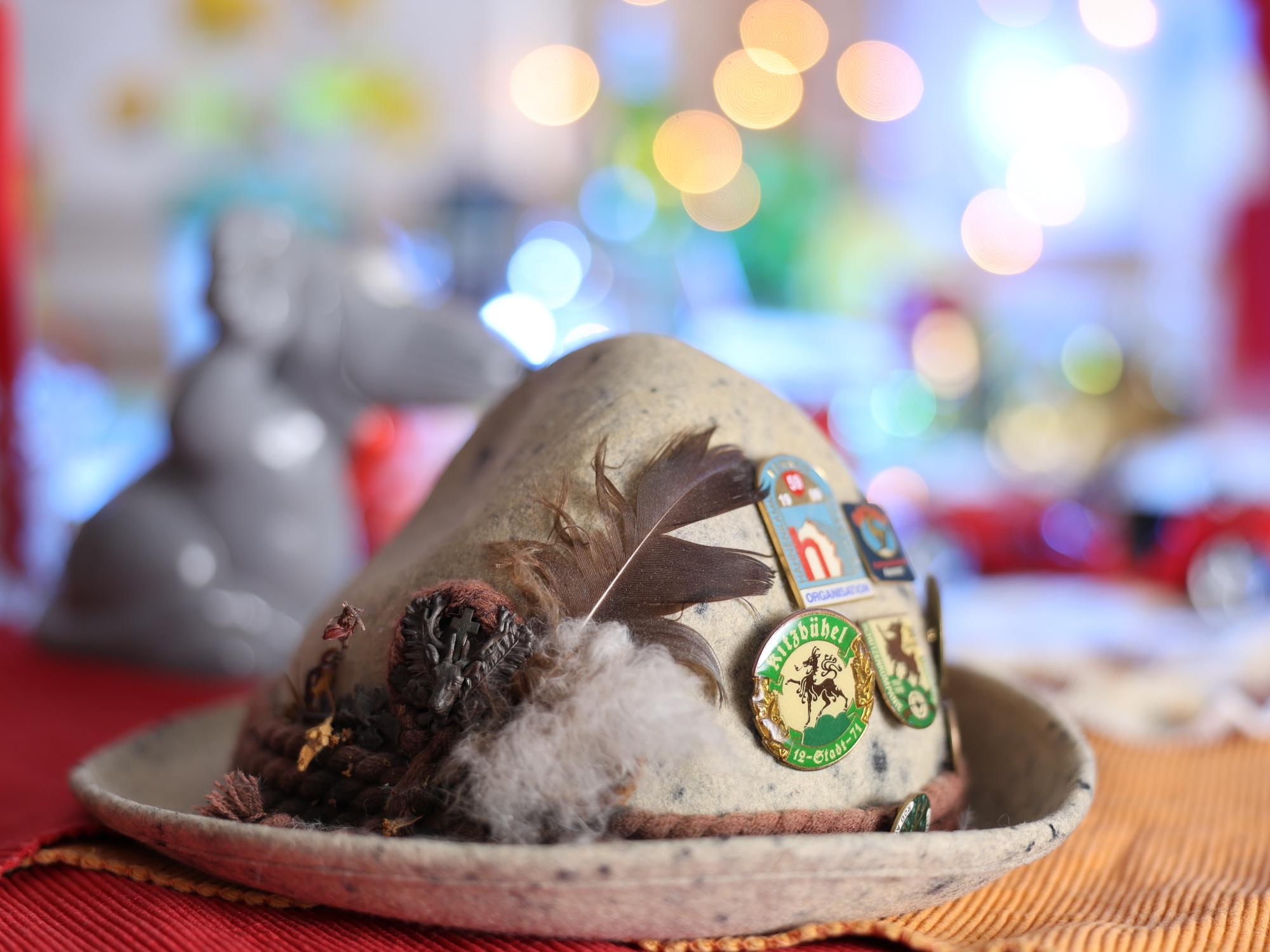} \\
    Sharp GT \fnum{22.0} & Blurry Input \fnum{7.1} & Blurry Input \fnum{4.0} & Blurry Input \fnum{2.8} & Blurry Input \fnum{2.0}  \\
    \end{tabular}
  \vspace{-2mm}
  \caption{\textbf{Representative samples from the RealDefocus~\cite{seizinger2025bokehlicious} dataset illustrating increasing defocus blur with decreasing aperture values.} For each scene, the image captured at \fnum{22.0} serves as the sharp ground truth, while progressively smaller \fstop values (e.g. \fnum{8.0}, \fnum{4.0} and \fnum{2.0}) produce stronger defocus blur effects. Two example scenes (outdoor and indoor) are shown at focal lengths of 61 and 70\,mm, respectively, highlighting the diversity in scene content, focal length, and blur strength.}
  \label{fig:RealDefocusExample}
\end{figure*}

\section{Introduction}
\label{sec:intro}
When capturing a scene with a finite depth of field, objects outside the focal plane appear blurred due to optical defocus. This phenomenon, commonly observed when using large aperture settings or imaging scenes with substantial depth variation, is often associated with the \textit{bokeh effect}~\cite{seizinger2025bokehlicious}. While shallow depth of field can be intentionally exploited for artistic purposes~\cite{kennerdell1997bokeh}, unintended defocus or misfocus frequently degrades image quality and hampers downstream vision tasks such as text recognition~\cite{wang2012end, chen2021text}, object detection~\cite{redmon2016you, molloy2024analysis}, and image classification~\cite{zhou2017classification}.

Single Image Defocus Deblurring (SIDD) aims to restore an all-in-focus image from a single defocused observation (see \cref{fig:RealDefocusExample}). The problem is inherently ill-posed: defocus blur is spatially varying and depends on scene depth as well as lens characteristics. Each observed pixel can be interpreted as a weighted aggregation of neighboring latent pixels, where the weights are governed by an unknown, spatially varying point spread function (PSF). Accurately estimating and inverting these blur kernels remains a central challenge.

Early Defocus Deblurring methods relied on explicit blur kernel estimation followed by non-blind deconvolution~\cite{levin2007image, d2016non}. More recently, deep learning-based approaches~\cite{abuolaim2020defocus, abuolaim2021learning, ruan2021aifnet, quan2021gaussian, quan2023neumann} have demonstrated substantial improvements by learning end-to-end mappings from blurred to sharp images. For such data-driven methods, the quality, diversity, and realism of the underlying training data are essential. Consequently, several datasets~\cite{ruan2021aifnet, abuolaim2020defocus, chen2025quad} have been introduced to support supervised learning for Defocus Deblurring, each with different trade-offs in scale, realism, and blur diversity.

Defocus Deblurring is closely related to, yet fundamentally distinct from, Motion Deblurring. While motion blur arises from camera or object movement during exposure and often results in complex, trajectory-dependent kernels, defocus blur is primarily governed by optical geometry and scene depth, typically exhibiting disk-like or near-circular PSFs with spatial variation. At the same time, Defocus Deblurring shares a strong conceptual connection with Bokeh Rendering~\cite{seizinger2025bokehlicious}, which can be interpreted as the inverse task: instead of removing optical blur, Bokeh Rendering aims to synthesize realistic shallow depth-of-field effects from sharp imagery.

In this work, we build upon the recently introduced RealBokeh/RealDefocus data collection paradigm~\cite{seizinger2025bokehlicious}, which provides large-scale, high-resolution, and well-aligned real-world aperture pairs (see \cref{fig:RealDefocusExample}). Compared to prior datasets, it offers significantly greater scene diversity, wider aperture coverage, and stronger defocus, enabling a more realistic and challenging evaluation of modern high-capacity architectures.

\section{Related Work}
\label{sec:related}

\begin{table}[t]
\centering
\small
\renewcommand{\arraystretch}{1}
\setlength{\tabcolsep}{5pt}
\begin{tabular}{l|cccc}
\toprule
                    & \textbf{RealDefocus}~\cite{seizinger2025bokehlicious} & DPDD$_S$~\cite{abuolaim2020defocus}   & LFDOF~\cite{ruan2021aifnet}   \\ 
\midrule
\# Samples          & \textbf{23,000}                                       & 500                                   & 12,600                         \\
\# Scenes           & \textbf{4,400}                                         & 500                                   & 805                           \\
\# Train            & \textbf{20,500}                                       & 350                                   & 11,850                         \\
\# Validation       & \textbf{1,250}                                        & 74                                    & 750                           \\
\# Test             & \textbf{1,250}                                        & 76                                    & -                             \\
Apertures           & \textbf{\fnum{2.0} - \fnum{20.0}}                     & \fnum{4.0}                            & 3                             \\
Focal Length        & 28-70mm                                               & 24-105mm                              & $\sim$50mm                        \\
Resolution          & 6000$\times$4000                                      & 6720$\times$4480                      & 1008$\times$688               \\
Defocus Blur        & Real                                                  & Real                                  & Synthetic                     \\
\bottomrule
\end{tabular}
\vspace{-2mm}
\caption{
\textbf{Comparison of representative training datasets for SIDD.} RealDefocus~\cite{seizinger2025bokehlicious} provides the largest scale in terms of total samples and unique scenes, with substantially more images than prior datasets. It also features a wide aperture range and high-resolution images, whereas DPDD$_S$~\cite{abuolaim2020defocus} offers only a fixed aperture setting while the multi aperture views of LFDOF~\cite{ruan2021aifnet} are synthetically generated.
}
\label{tab:datasets}
\end{table}

\subsection{Defocus Deblurring Datasets}

One of the first datasets enabling quantitative evaluation was RTF~\cite{d2016non}, captured using a Lytro light field camera~\cite{ng2006digital}. While innovative, RTF contains only 22 images at a low spatial resolution ($360\times360$), with its defocus blur synthetically rendered from light field data. Consequently, its scale and realism limit its suitability for training modern deep nets.

DPDD~\cite{abuolaim2020defocus} marked a significant step forward as the first dataset explicitly designed for training deep Defocus Deblurring models. Captured with a Canon EOS 5D Mark IV, it leverages dual-pixel (DP) views to facilitate blur estimation. However, DP data requires specialized sensor hardware and is therefore not widely available. Moreover, DPDD$_S$ is relatively small in scale, restricted to a single aperture setting (\fnum{4.0}), and thus exhibits comparatively mild defocus blur.

LFDOF~\cite{ruan2021aifnet} increased dataset size and scene diversity by capturing light field data~\cite{dansereau2019liff}, allowing multiple focus planes and blur levels per scene. Nevertheless, the defocus blur is synthetically generated from the recorded light field, which introduces a noticeable domain gap to real optical blur~\cite{ruan2022learning}.

RealDOF~\cite{lee2021iterative} provides high-quality real defocus pairs captured with two DSLRs and a split-view mirror system. While offering good alignment and a resolution of $2320\times1536$, it contains only 50 images and serves primarily as an evaluation set rather than a training resource.

\subsection{RealDefocus}

The \textit{RealDefocus} dataset~\cite{seizinger2025bokehlicious} originally proposed by Seizinger \etal, addresses the limitations of prior datasets in terms of scale, diversity, and aperture coverage. It is derived from the \textit{RealBokeh} dataset for Bokeh Rendering, which captures multiple aperture settings per scene. By reversing the roles of input and ground truth, which means using shallow depth-of-field images as inputs and the image captured at \fnum{22.0} as the sharp reference, RealDefocus naturally forms paired data for \task (\stask) as shown in \cref{fig:RealDefocusExample}. 

As summarized in \cref{tab:datasets}, RealDefocus contains 23,000 image pairs across 4,400 unique scenes, significantly surpassing previous real-world SIDD datasets in scale. Images are captured at a high resolution of $6000\times4000$ and span a wide aperture range from \fnum{2.0} to \fnum{20.0}, enabling evaluation across varying blur strengths.

The dataset includes diverse indoor and outdoor environments, controlled studio scenes with complex highlight structures, portraits, and challenging night-time scenes, conditions largely underrepresented in earlier datasets.
Importantly, the maximum aperture of \fnum{2.0} produces substantially stronger defocus blur than the \fnum{4.0} setting used in DPDD$_S$, resulting in a more challenging and realistic benchmark. By combining large scale, high resolution, wide aperture variability, and real optical blur, RealDefocus establishes a new foundation for training and evaluating modern \stask~methods.

\begin{table*}[t]
\centering
\small
\setlength{\tabcolsep}{1.4pt}
\begin{tabular}{l|ccc|ccc|ccc|ccc|ccc}
\toprule
  \multirow{2}{*}{Method} &
  Param. & MACs & \phantom{.}Time\phantom{.} &
  \multicolumn{3}{c|}{\fnum{2.0}} &
  \multicolumn{3}{c|}{\fnum{4.0}} &
  \multicolumn{3}{c|}{\fnum{8.0}} &
  \multicolumn{3}{c}{all \fstops} \\
  \addlinespace[0.5pt]
 &
% parameters
    M.$\downarrow$ &
% macs
    G.$\downarrow$ &
% runtime
    sec$\downarrow$ &
%F2.0
    PSNR$\uparrow$ &
    SSIM$\uparrow$ &
    LPIPS$\downarrow$ &
%F4.0
    PSNR$\uparrow$ &
    SSIM$\uparrow$ &
    LPIPS$\downarrow$ &
% %F8.0
    PSNR$\uparrow$ &
    SSIM$\uparrow$ &
    LPIPS$\downarrow$ &
% avg
    PSNR$\uparrow$ &
    SSIM$\uparrow$ &
    LPIPS$\downarrow$ \\ 
    \midrule
Input &
% parameters
    - &
% macs
    - &
% Runtime
    - &
%F2.0
    19.666 & % PSNR
    0.6434 & % SSIM
    0.3238 & % LPIPS
%4.0
    21.324 & % PSNR
    0.7137 & % SSIM
    0.4028 & % LPIPS
 %8.0
    25.780 & % PSNR
    0.8084 & % SSIM
    0.2403 & % LPIPS
% Avg
    24.422 & % PSNR
    0.7668 & % SSIM
    0.3237 \\ %LPIPS
\midrule
% Method &
% % parameters
%     - &
% % macs
%     - &
% % Runtime
%     - &
% %F2.0
%     - & % PSNR
%     - & % SSIM
%     - & % LPIPS
% %4.0
%     - & % PSNR
%     - & % SSIM
%     - & % LPIPS
%  %8.0
%     - & % PSNR
%     - & % SSIM
%     - & % LPIPS
% % Avg
%     - & % PSNR
%     - & % SSIM
%     - \\ %LPIPS
DPDNet$_S$~\cite{abuolaim2020defocus} &
% parameters
    31.38 &
% macs
    \phantom{0}\textit{55.92} &
% Runtime
    \textbf{0.043} &
%F2.0
    21.112 & % PSNR
    0.6643 & % SSIM
    0.4584 & % LPIPS
%4.0
    23.947 & % PSNR
    0.7626 & % SSIM
    0.3359 & % LPIPS
 %8.0
    28.332 & % PSNR
    0.8601 & % SSIM
    0.1705 & % LPIPS
% Avg
    25.887 & % PSNR
    0.7946 & % SSIM
    0.2768 \\ %LPIPS
\rowcolor{lgray}
GKMNet~\cite{quan2021gaussian} &
% parameters
    \phantom{0}\textbf{1.41} &
% macs
    \phantom{0}\textbf{20.05} &
% Runtime
    0.135 &
%F2.0
    21.946 & % PSNR
    0.6821 & % SSIM
    0.4220 & %LPIPS
%4.0
    25.523 & % PSNR
    0.7951 & % SSIM
    0.2748 & % LPIPS
 %8.0
    29.400 & % PSNR
    0.8769 & % SSIM
    0.1438 & % LPIPS
% Avg
    27.166 & % PSNR
    0.8157 & % SSIM
    0.2393 \\% LPIPS
DRBNet~\cite{ruan2022learning} &
% parameters
    \textit{11.69} &
% macs
    \phantom{0}\underline{44.47} &
% Runtime
    \underline{0.068} &
%F2.0
    22.162 & % PSNR
    0.7011 & % SSIM
    0.3626 & % LPIPS
%4.0
    26.193 & % PSNR
    0.8220 & % SSIM
    0.2126 & % LPIPS
 %8.0
    30.639 & % PSNR
    0.9010 & % SSIM
    0.1016 & % LPIPS
% Avg
    28.022 & % PSNR
    0.8374 & % SSIM
    0.1910 \\ %LPIPS
\rowcolor{lgray}
NRKNet~\cite{quan2023neumann} &
% parameters
    \phantom{0}\underline{6.09} &
% macs
    \phantom{0}61.81 &
% Runtime
    \textit{0.096} &
%F2.0
    22.581 & % PSNR
    0.7160 & % SSIM
    0.4042 & % LPIPS
%4.0
    26.542 & % PSNR
    0.8332 & % SSIM
    0.2450 & % LPIPS
 %8.0
    30.702 & % PSNR
    0.9075 & % SSIM
    0.1218 & % LPIPS
% Avg
    28.211 & % PSNR
    0.8461 & % SSIM
    0.2195 \\ %LPIPS
% \rowcolor{lgray}
Restormer~\cite{zamir2022restormer} &
% parameters
    26.13 &
% macs
    141.24 &
% Runtime
    0.758 &
%F2.0
    21.947 & % PSNR
    0.7033 & % SSIM
    0.3500 & % LPIPS
%4.0
    27.226 & % PSNR
    0.8483 & % SSIM
    0.1986 & % LPIPS
 %8.0
    31.726 & % PSNR
    0.9211 & % SSIM
    0.0942 & % LPIPS
% Avg
    28.844 & % PSNR
    0.8538 & % SSIM
    0.1802 \\ %LPIPS
\rowcolor{lgray}
EVSSM~\cite{kong2025efficient} &
% parameters
    17.13 &
% macs
    122.30 &
% Runtime
    1.398 &
%F2.0
    22.891 & % PSNR
    0.7276 & % SSIM
    0.3771 & % LPIPS
%4.0
    27.096 & % PSNR
    0.8466 & % SSIM
    0.2137 & % LPIPS
 %8.0
    31.761 & % PSNR
    0.9214 & % SSIM
    0.0985 & % LPIPS
% Avg
    29.052 & % PSNR
    0.8592 & % SSIM
    0.1934 \\ %LPIPS
% \rowcolor{lgray}
LAKDNet~\cite{ruan2023revisiting} &
% parameters
    17.73 &
% macs
    \phantom{0}86.52 &
% Runtime
    0.803 &
%F2.0
    23.182 & % PSNR
    0.7360 & % SSIM
    %% \textit{0.2860} & % LPIPS
    \underline{0.2860} & % LPIPS
%4.0
    27.348 & % PSNR
    0.8478 & % SSIM
    \underline{0.1598} & % LPIPS
 %8.0
    31.858 & % PSNR
    0.9201 & % SSIM
    \underline{0.0774} & % LPIPS
% Avg
    29.156 & % PSNR
    0.8611 & % SSIM
    \underline{0.1474} \\ %LPIPS
\rowcolor{lgray}
EAMamba~\cite{lin2025eamamba} &
% parameters
    25.30 &
% macs
    \phantom{0}84.29 &
% Runtime
    1.216 &
%F2.0
    \textit{23.552} & % PSNR
    \textit{0.7460} & % SSIM
    0.3240 & % LPIPS
%4.0
    \textit{27.694} & % PSNR
    \textit{0.8543} & % SSIM
    0.1898 & % LPIPS
 %8.0
    \underline{32.080} & % PSNR
    \underline{0.9239} & % SSIM
    0.0908 & % LPIPS
% Avg
    \textit{29.475} & % PSNR
    \textit{0.8672} & % SSIM
    0.1695 \\ %LPIPS
% \rowcolor{lgray}
Bokehlicious~\cite{seizinger2025bokehlicious} &
% parameters
    13.96 &
% macs
    \phantom{0}60.55 &
% Runtime
    0.523 &
%F2.0
    \textbf{24.546} & % PSNR
    \underline{0.7732} & % SSIM
    \textbf{0.2043} & % LPIPS
%4.0
    \underline{28.141}& % PSNR
    \underline{0.8607} & % SSIM
    \textbf{0.1275} & % LPIPS
 %8.0
    \textit{31.975} & % PSNR
    \textit{0.9219} & % SSIM
    \textbf{0.0676} & % LPIPS
% Avg
    \underline{29.752} & % PSNR
    \underline{0.8751} & % SSIM
    \textbf{0.1125} \\ %LPIPS
    %FID 42.46
\rowcolor{lgray}
FFTFormer~\cite{kong2023efficient} &
% parameters
    16.56 &
% macs
    131.75 &
% Runtime
    2.232 &
%F2.0
    \underline{24.522} & % PSNR
    \textbf{0.7799} & % SSIM
    %% \underline{0.2991} & % LPIPS
    \textit{0.2991} & % LPIPS
%4.0
    \textbf{28.459} & % PSNR
    \textbf{0.8719} & % SSIM
    \textit{0.1670} & % LPIPS
 %8.0
    \textbf{32.818} & % PSNR
    \textbf{0.9331} & % SSIM
    \textit{0.0810} & % LPIPS
% Avg
    \textbf{30.206} & % PSNR
    \textbf{0.8837} & % SSIM
    \textit{0.1544} \\ %LPIPS
\bottomrule
\end{tabular}
\vspace{-2mm}
\caption{\textbf{Quantitative results on the RealDefocus~\cite{seizinger2025bokehlicious} SIDD benchmark.} Columns labeled by \fstop~values report performance when the input is restricted to the corresponding aperture setting; smaller \fstop~values indicate stronger defocus blur. The last column shows the average performance over the entire RealDefocus dataset. The best, second-best, and third-best results are highlighted in \textbf{bold}, \underline{underline}, and \textit{italic}, respectively. Computational cost in Giga-Multiply-Accumulate-Operations (GMACs) is reported at 256$\times$256 resolution, and inference time is measured at 1024$\times$1024 on an NVIDIA L40 GPU.
}
\label{tab:RealDefocusBench}
\end{table*}

\section{Benchmark Methodology}
\label{sec:methodology}

\begin{figure*}[t]
\begin{center}
\renewcommand{\arraystretch}{0.4}
\setlength{\tabcolsep}{0.8pt}
\footnotesize
\def\imgcompp{.176}
\def\imgcomp{.172}
\def\widthcompp{.14}
\begin{tabular}[b]{c@{ } c@{ }  c@{ } c@{ } c@{ }  c@{ }}\hspace{-4mm}
\fnum{2.0} Input &  GT: PSNR (dB) & DRBNet: 25.28 & NRKNet: 25.47 & Restormer: 24.79 & EVSSM: 25.90 \\
\addlinespace[0.75pt]
\multirow{3}{*}[1.6mm]{\includegraphics[height=\imgcompp\textwidth, trim={200px 100px 300px 200px},clip]{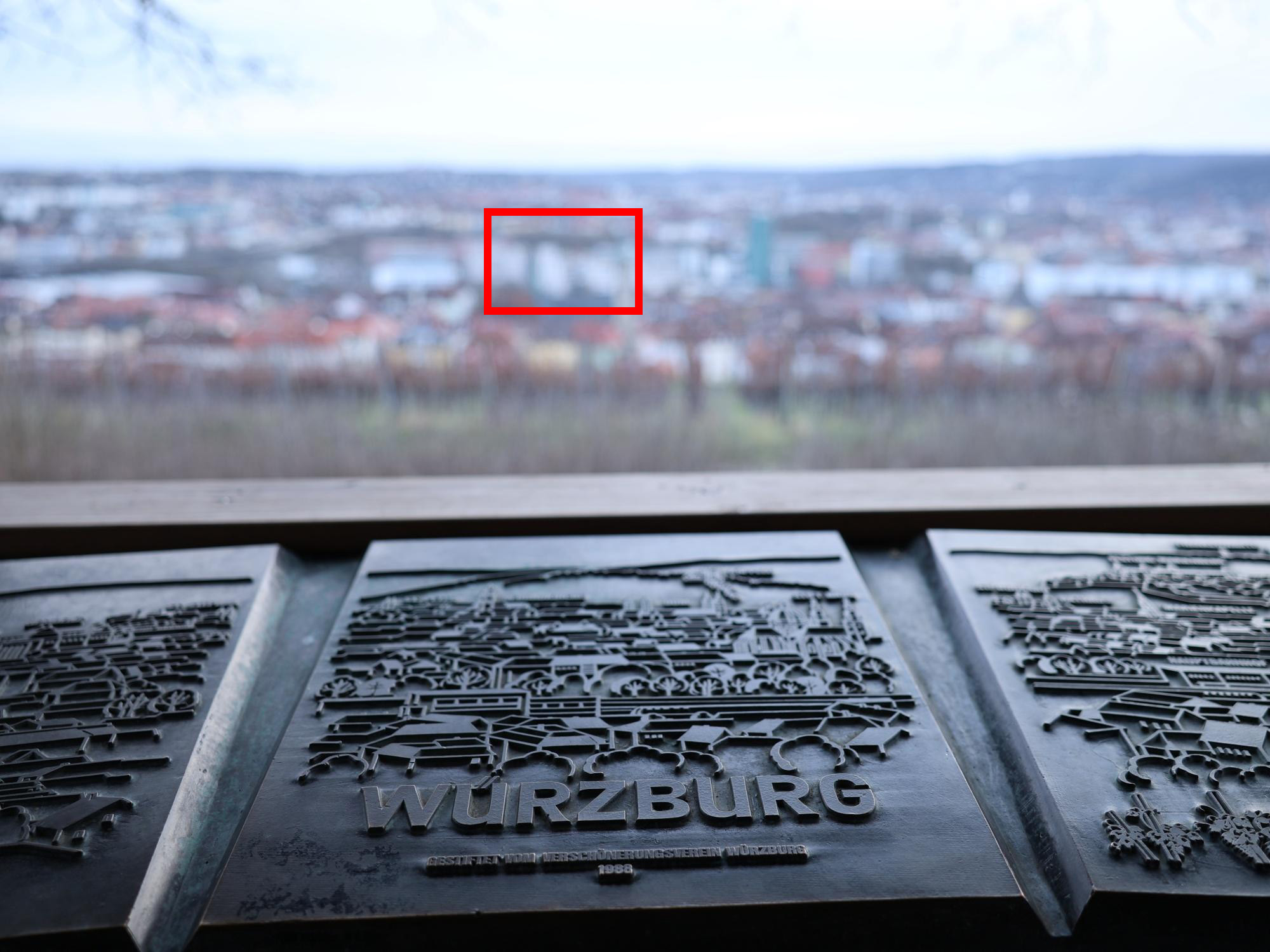}} &
            \includegraphics[width=\widthcompp\textwidth,valign=t, trim={770px 1020px 1000px 340px},clip]{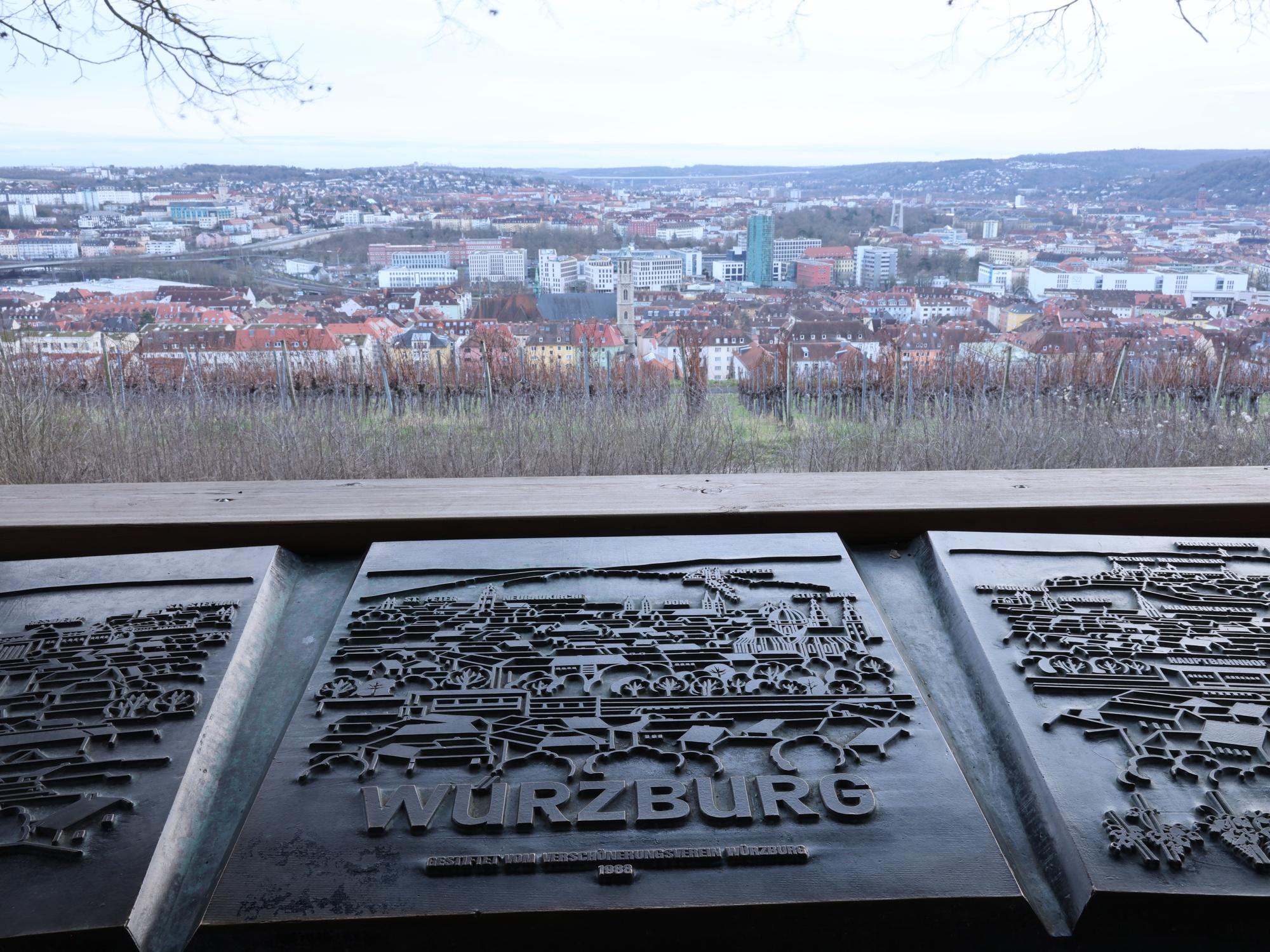} &
            \includegraphics[width=\widthcompp\textwidth,valign=t, trim={770px 1016px 1000px 332px},clip]{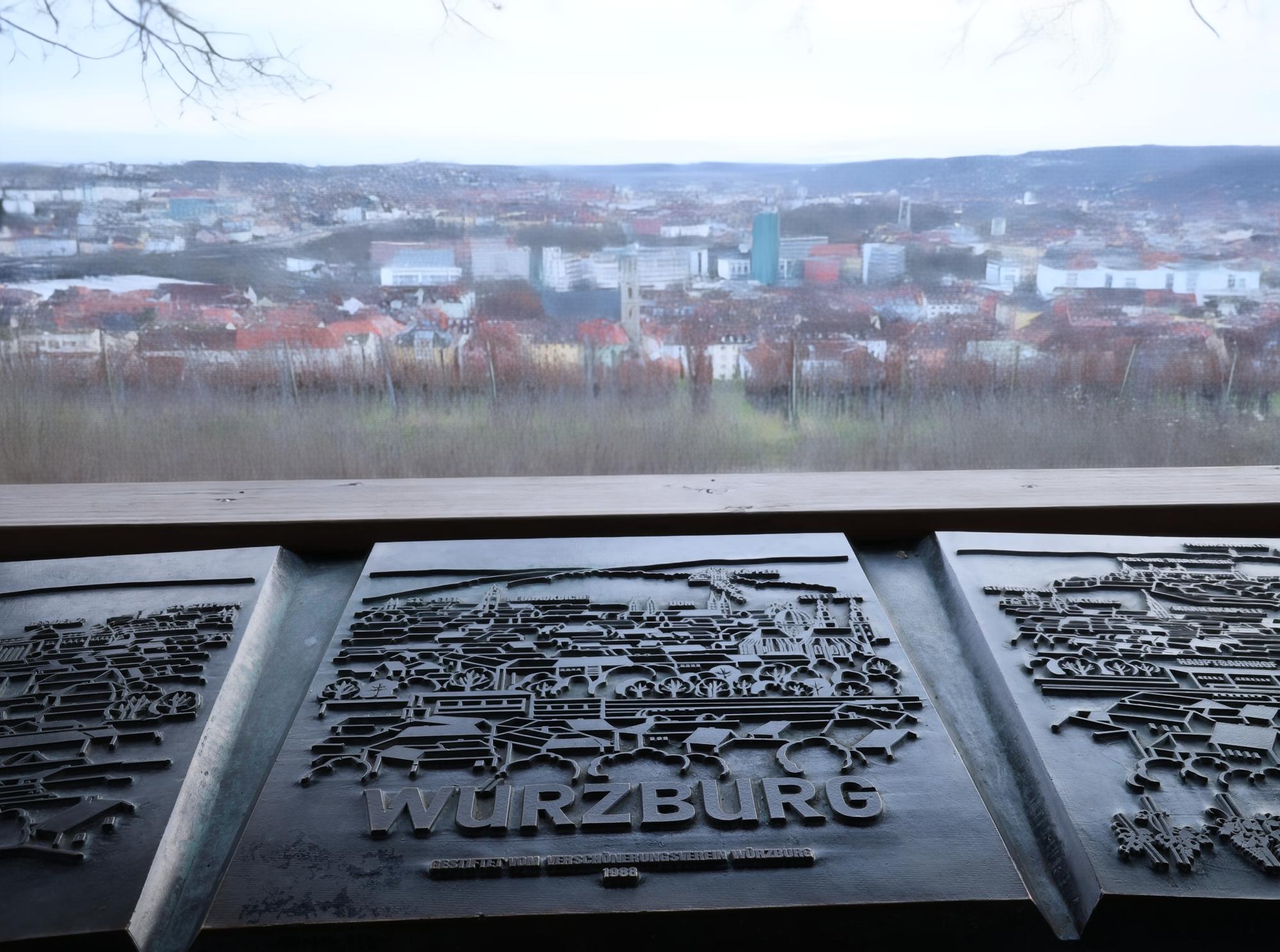} &
            \includegraphics[width=\widthcompp\textwidth,valign=t, trim={764px 1004px 984px 324px},clip]{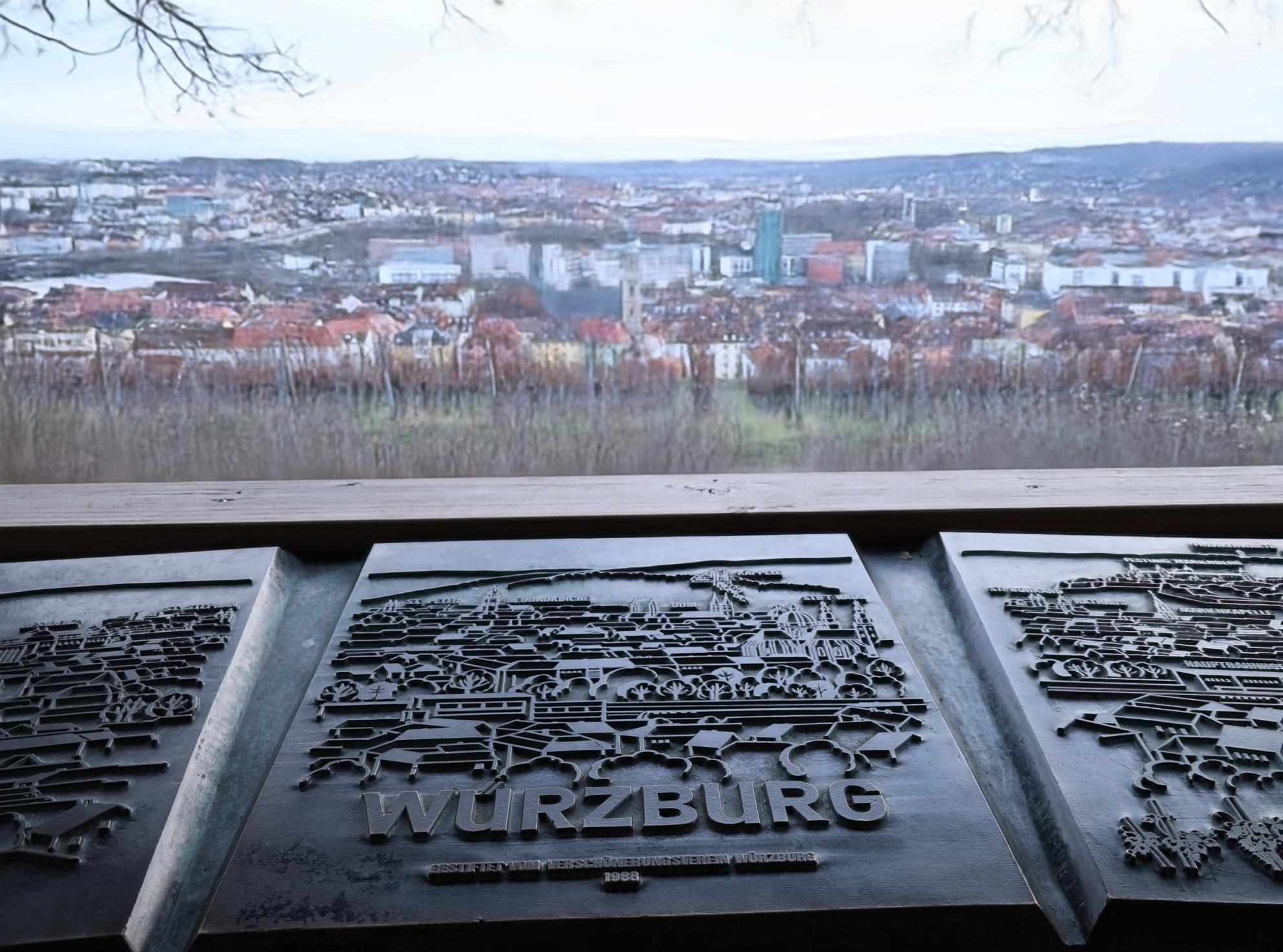} &
            \includegraphics[width=\widthcompp\textwidth,valign=t, trim={770px 1016px 1000px 332px},clip]{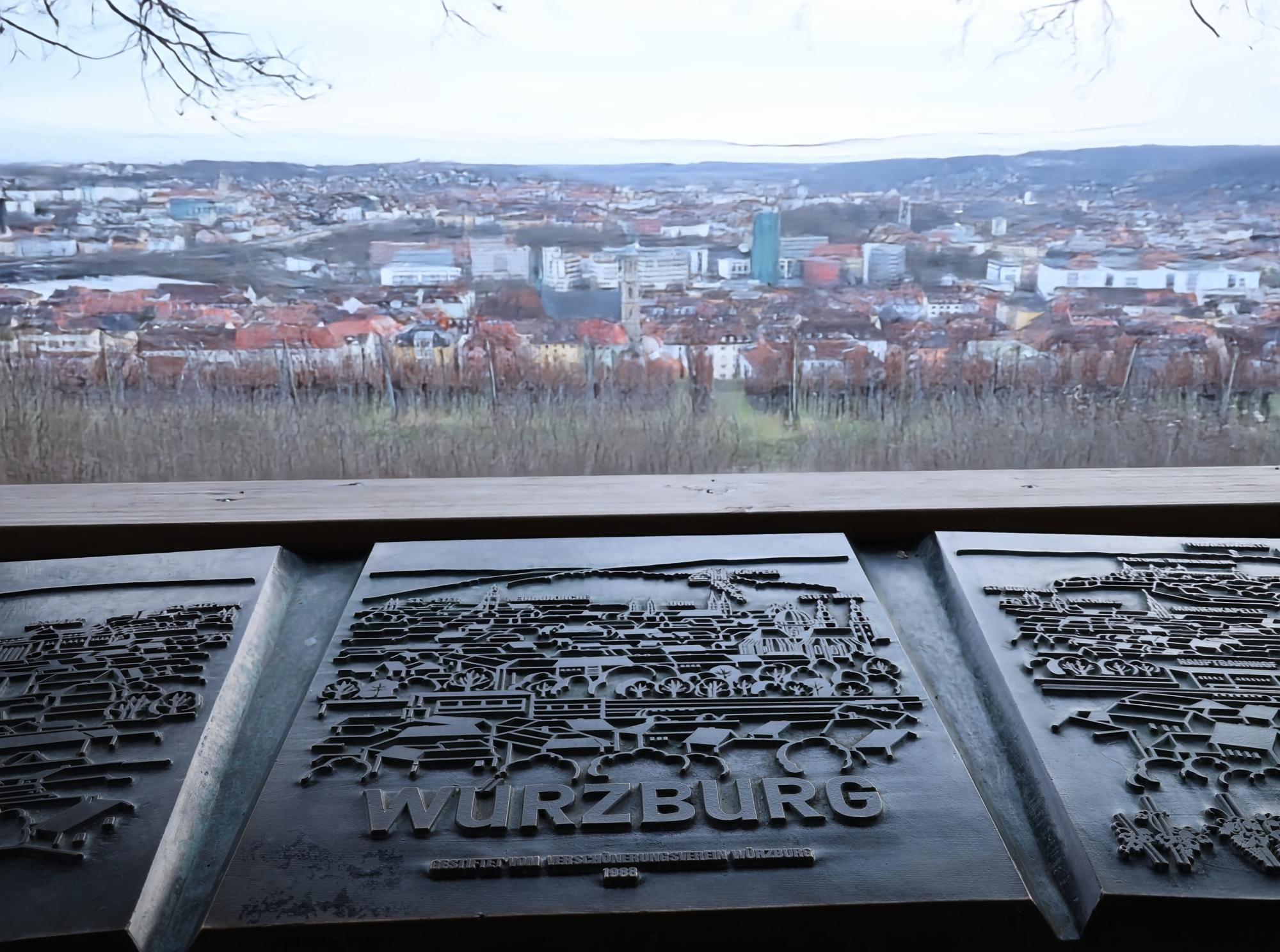} &
            \includegraphics[width=\widthcompp\textwidth,valign=t, trim={764px 1004px 984px 324px},clip]{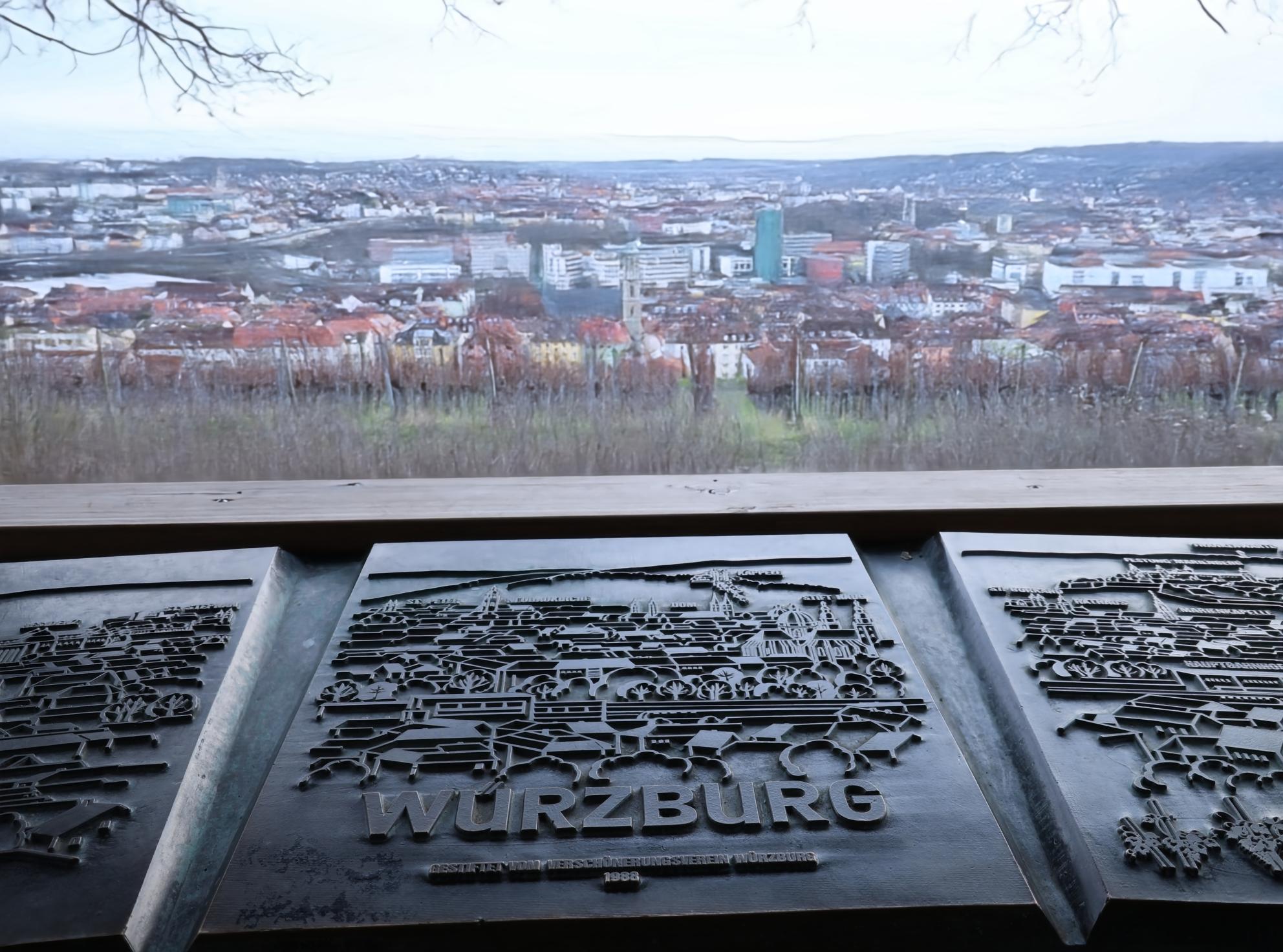}
            \\
            \addlinespace[2.0pt]
            &
            \includegraphics[width=\widthcompp\textwidth,valign=t, trim={770px 1020px 1000px 340px},clip]{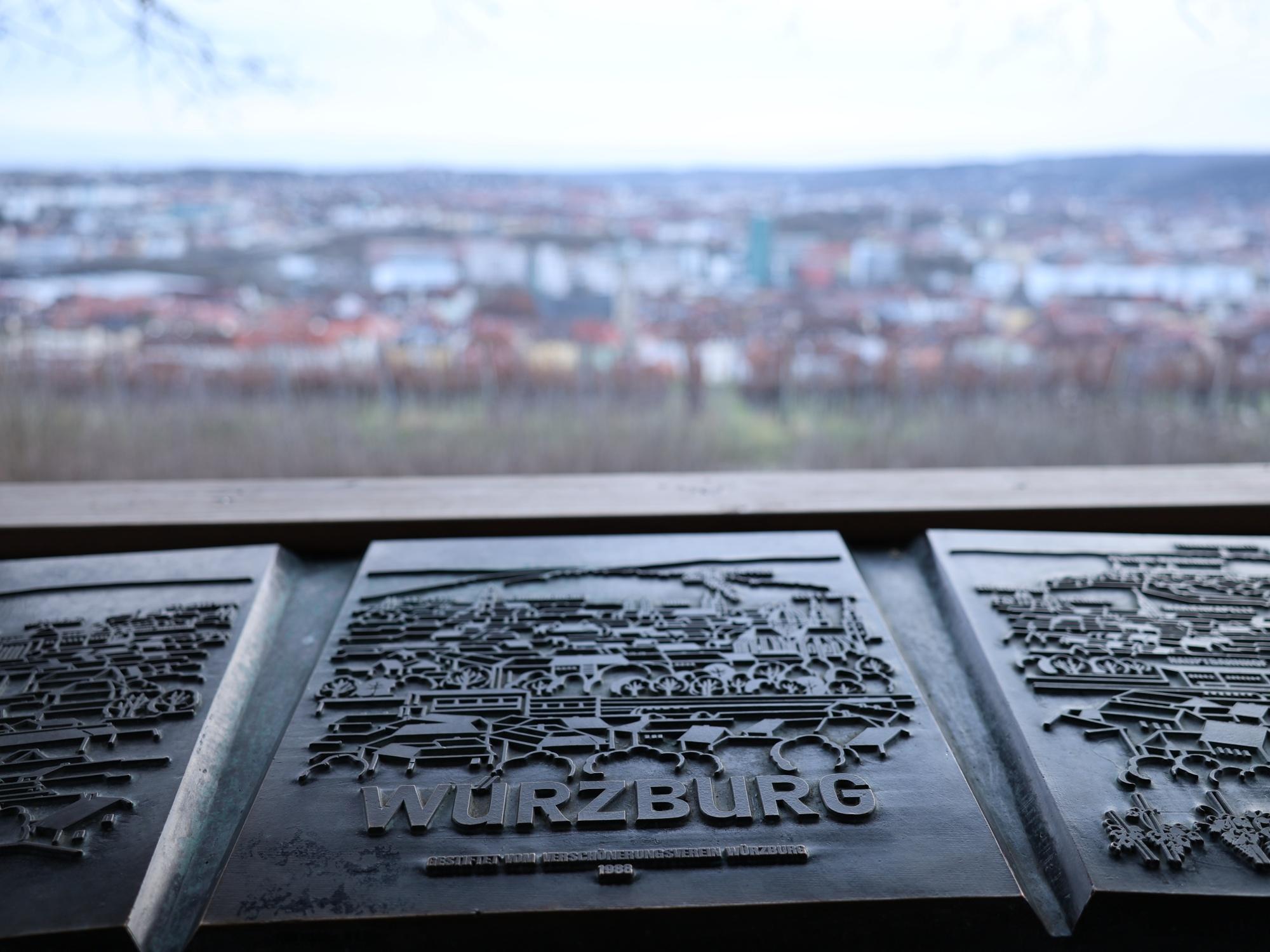} &
            \includegraphics[width=\widthcompp\textwidth,valign=t, trim={770px 1016px 1000px 332px},clip]{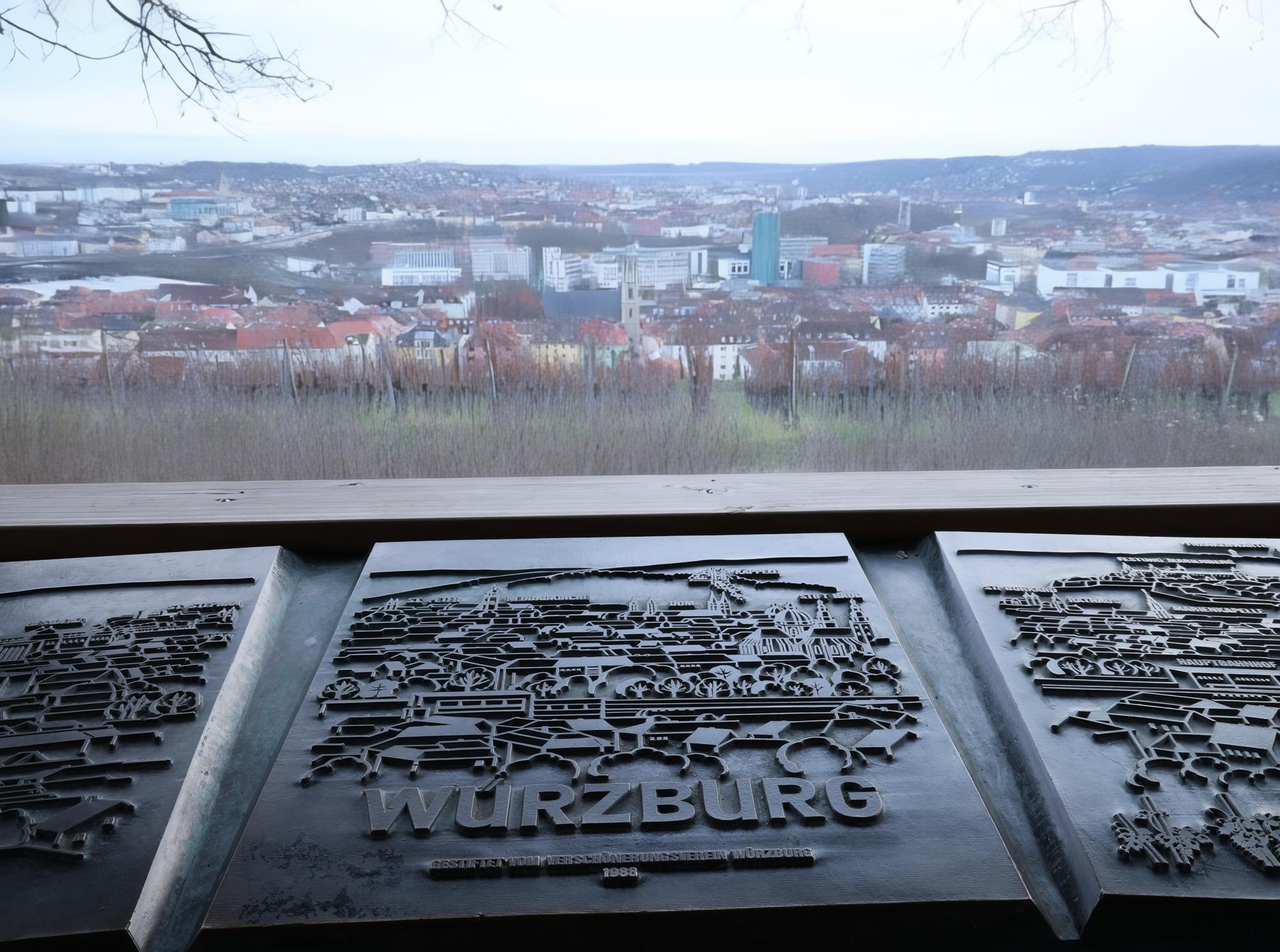} &
            \includegraphics[width=\widthcompp\textwidth,valign=t, trim={770px 1016px 1000px 332px},clip]{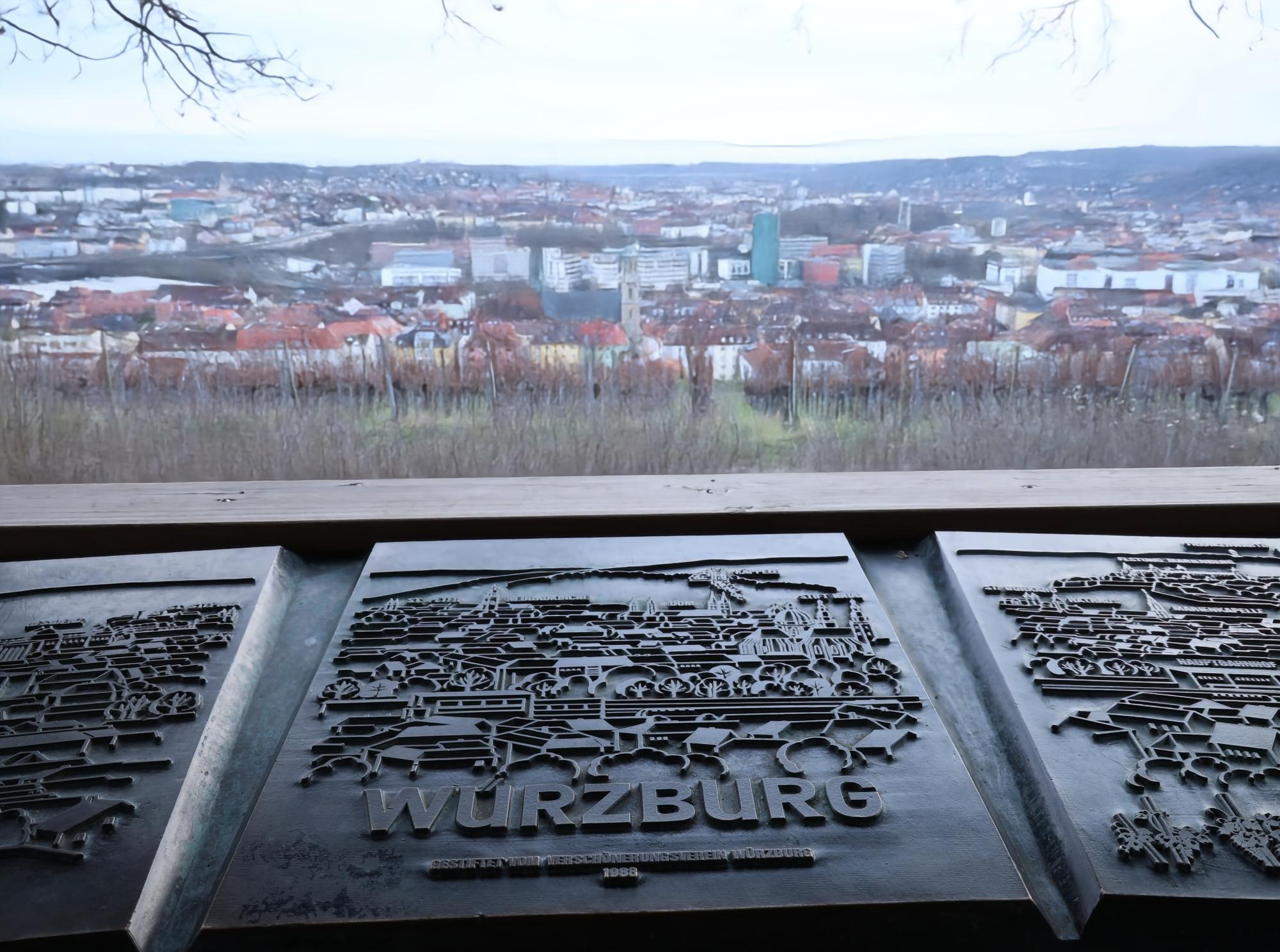} &
            \includegraphics[width=\widthcompp\textwidth,valign=t, trim={770px 1020px 1000px 340px},clip]{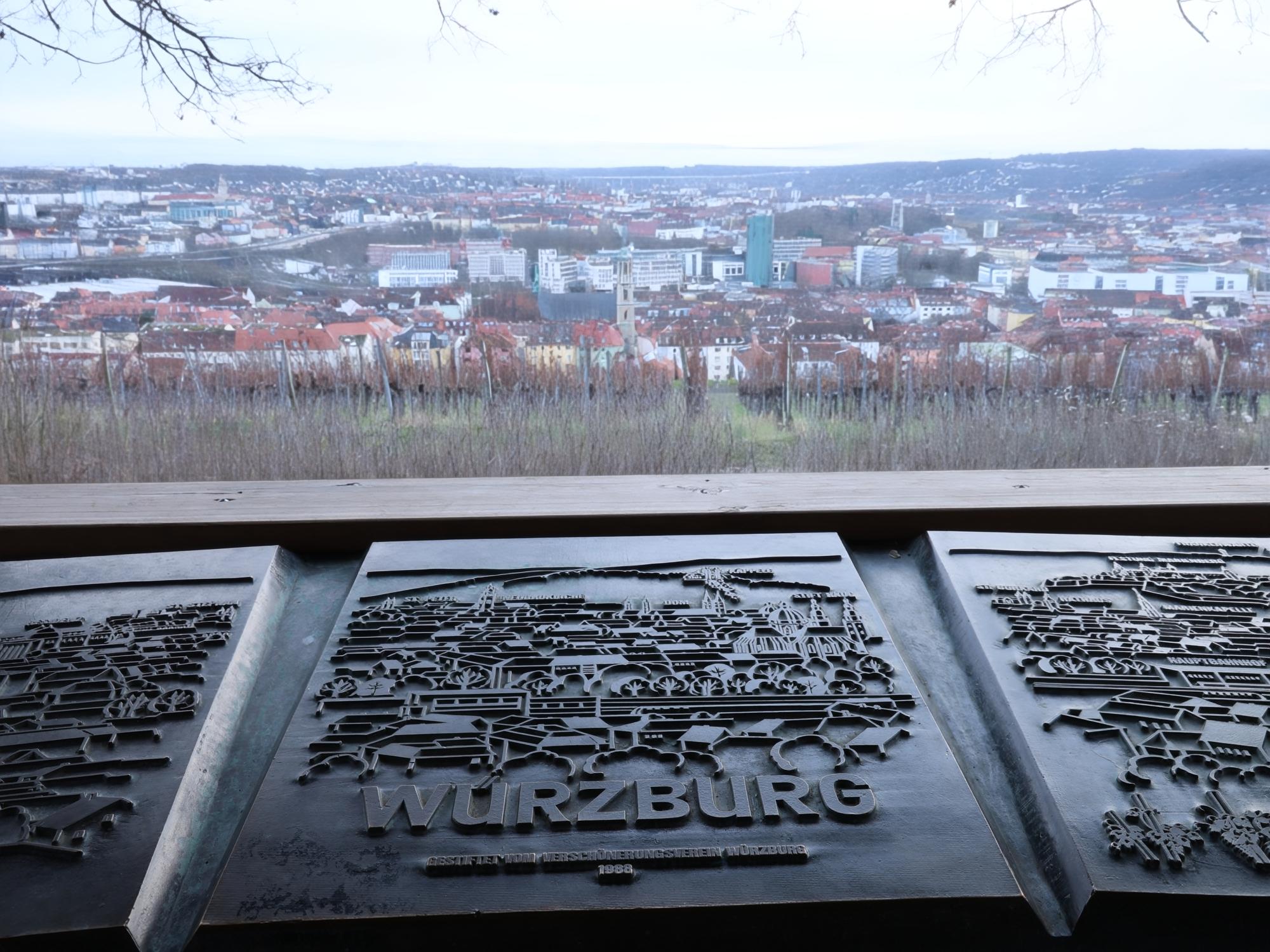} &
            \includegraphics[width=\widthcompp\textwidth,valign=t, trim={764px 1004px 984px 324px},clip]{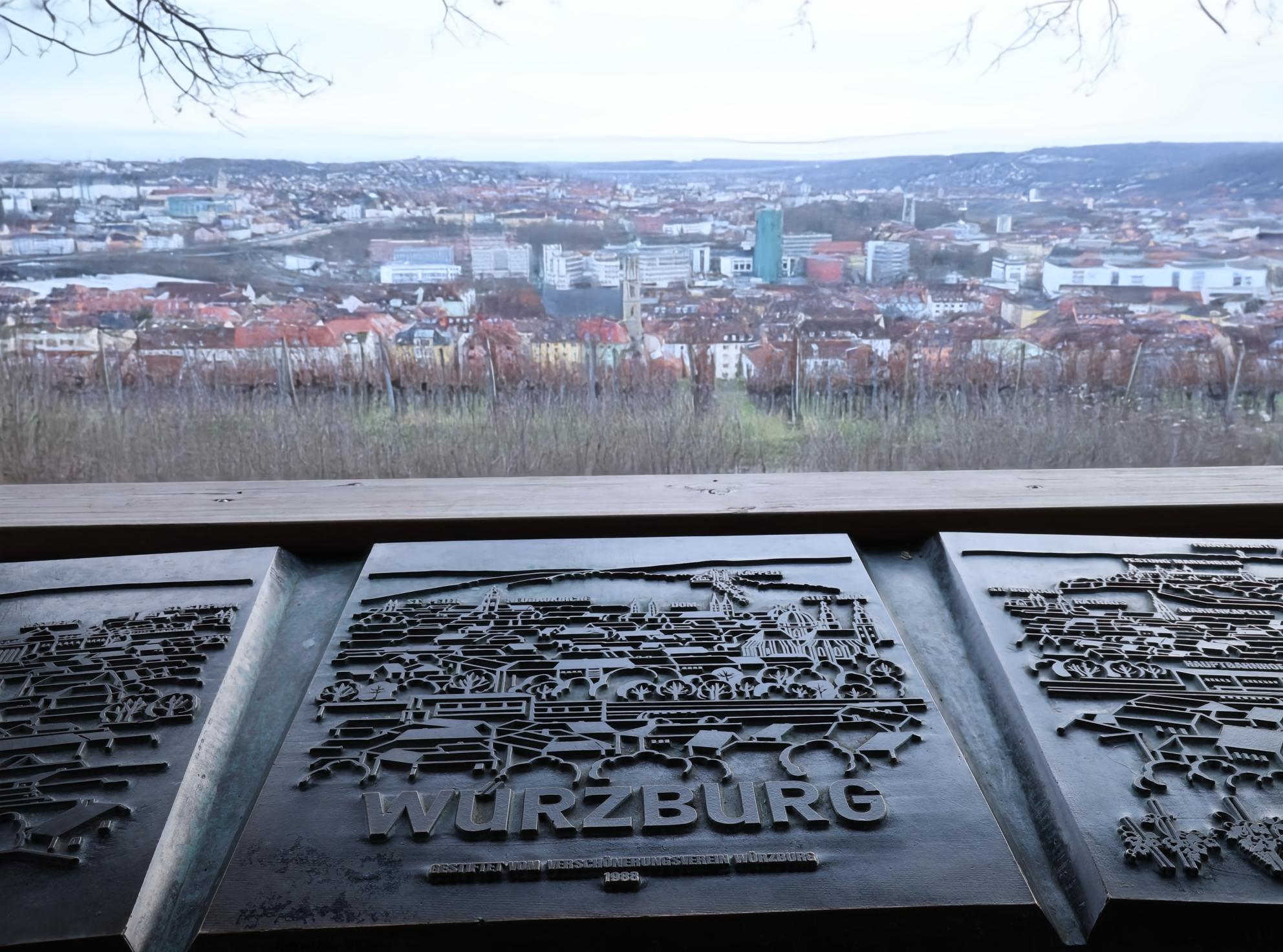}
            \\
            \addlinespace[1pt]
            & \fnum{2.0} Input: 21.81 & LAKDNet: 26.46 & EAMamba: 26.85 & Bokehlicious: 27.24 & FFTFormer: 27.69 \\
\addlinespace[3.0pt] % Next Aperture %%%%%%%%%%%%%%%%%%%%%%%%%%%%%%%%%%%%%%%%%%%%%%%%%%%%%%%%%%%%%%%%%%%%%%%%%5
\fnum{2.8} Input &  GT: PSNR (dB) & DRBNet: 24.11 & NRKNet: 24.16 & Restormer: 24.34 & EVSSM: 24.51 \\
\addlinespace[0.75pt]
\multirow{3}{*}[1.42mm]{\includegraphics[height=\imgcomp\textwidth, trim={180px 150px 280px 150px},clip]{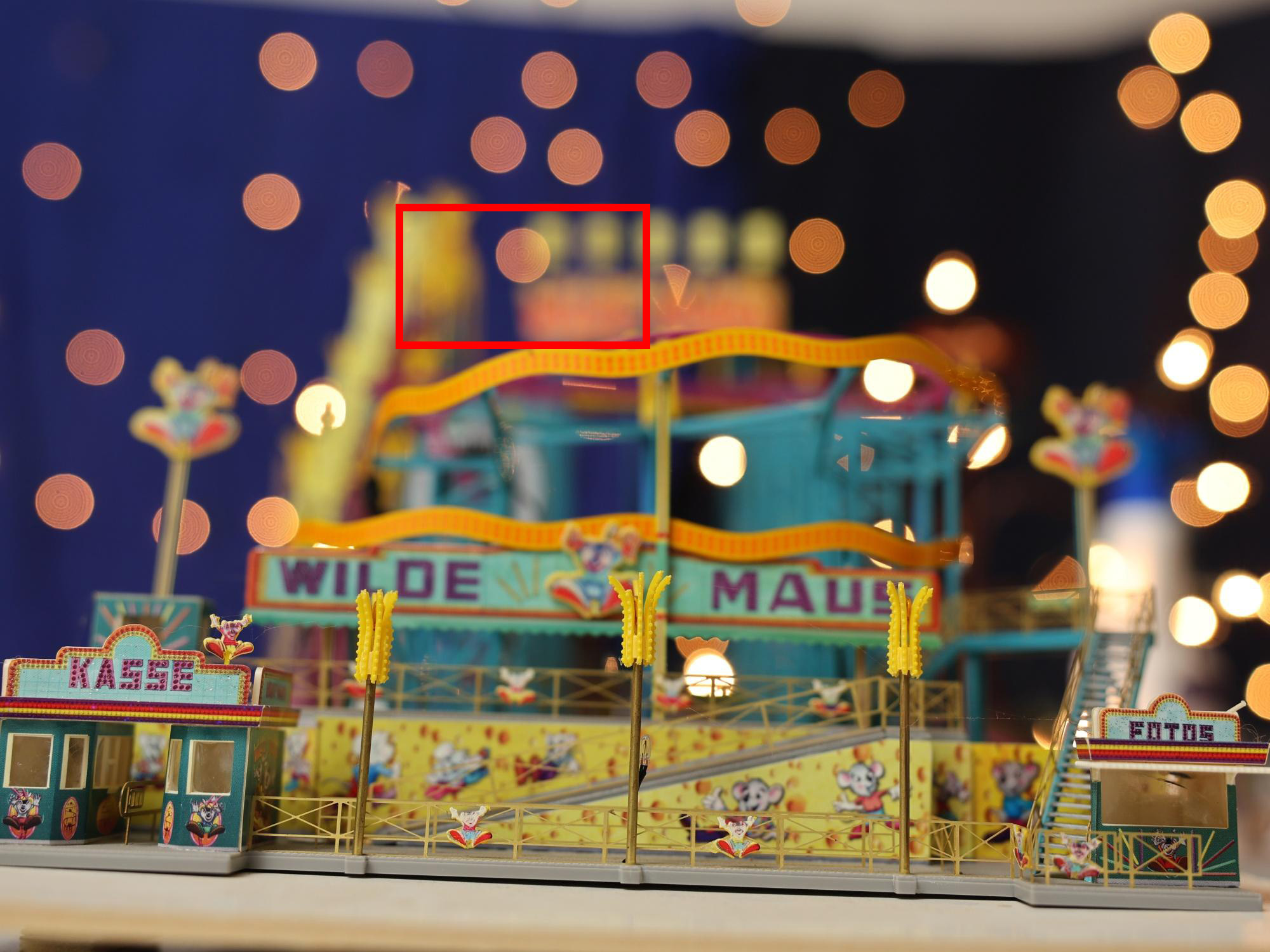}} &
            \includegraphics[width=\widthcompp\textwidth,valign=t, trim={650px 950px 1000px 340px},clip]{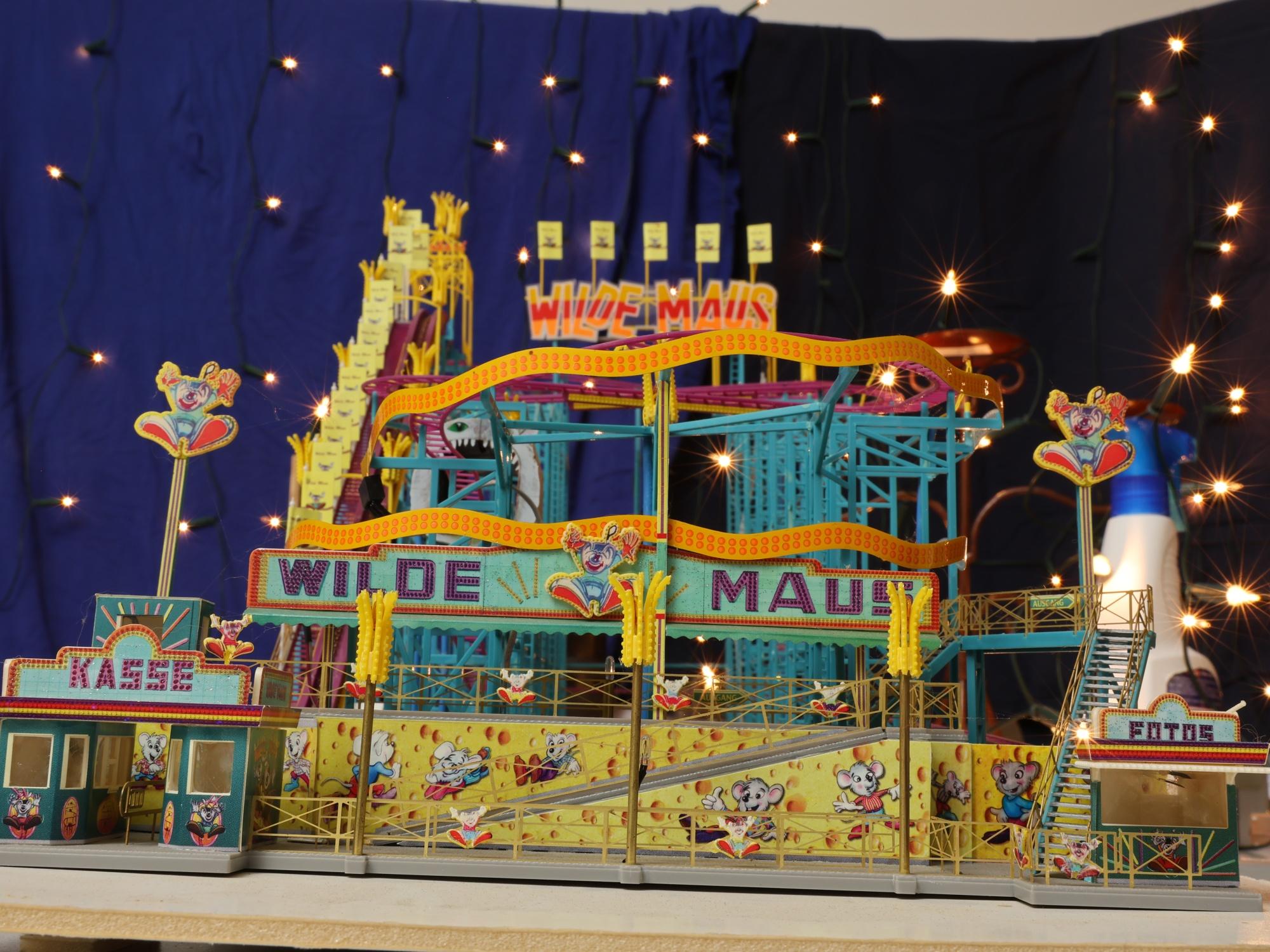} &
            \includegraphics[width=\widthcompp\textwidth,valign=t, trim={650px 944px 1000px 334px},clip]{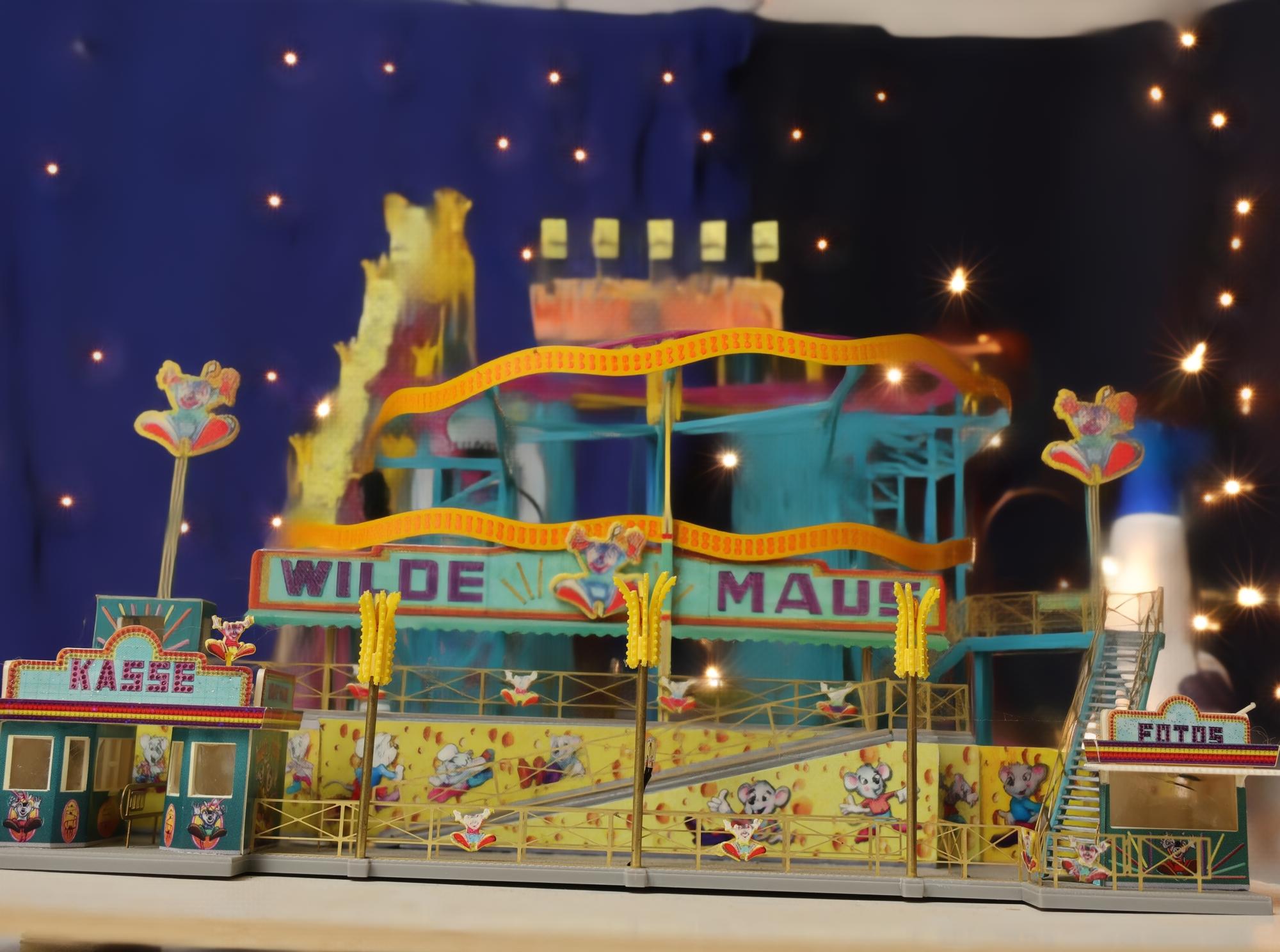} &
            \includegraphics[width=\widthcompp\textwidth,valign=t, trim={650px 940px 1000px 332px},clip]{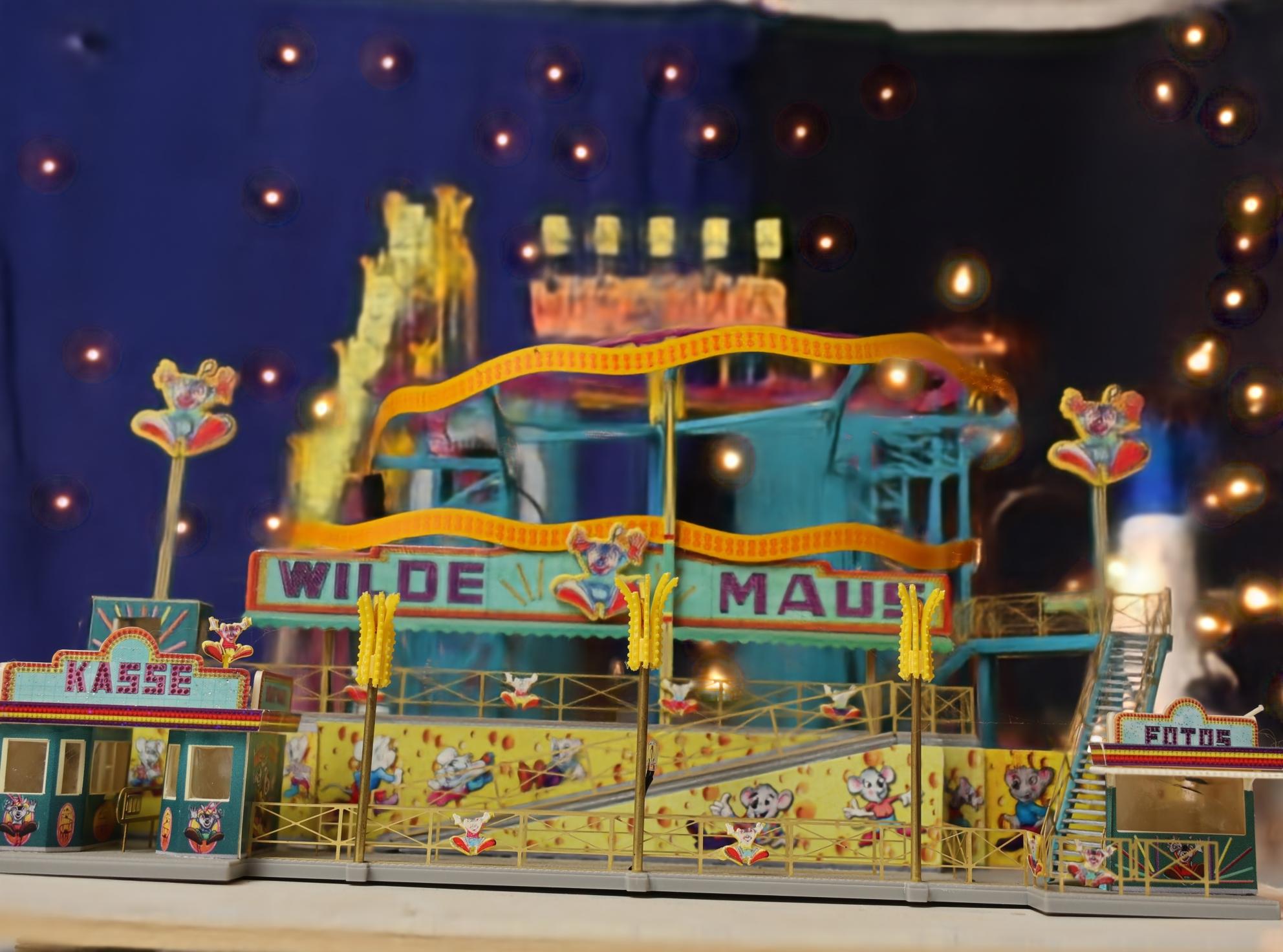} &
            \includegraphics[width=\widthcompp\textwidth,valign=t, trim={650px 944px 1000px 334px},clip]{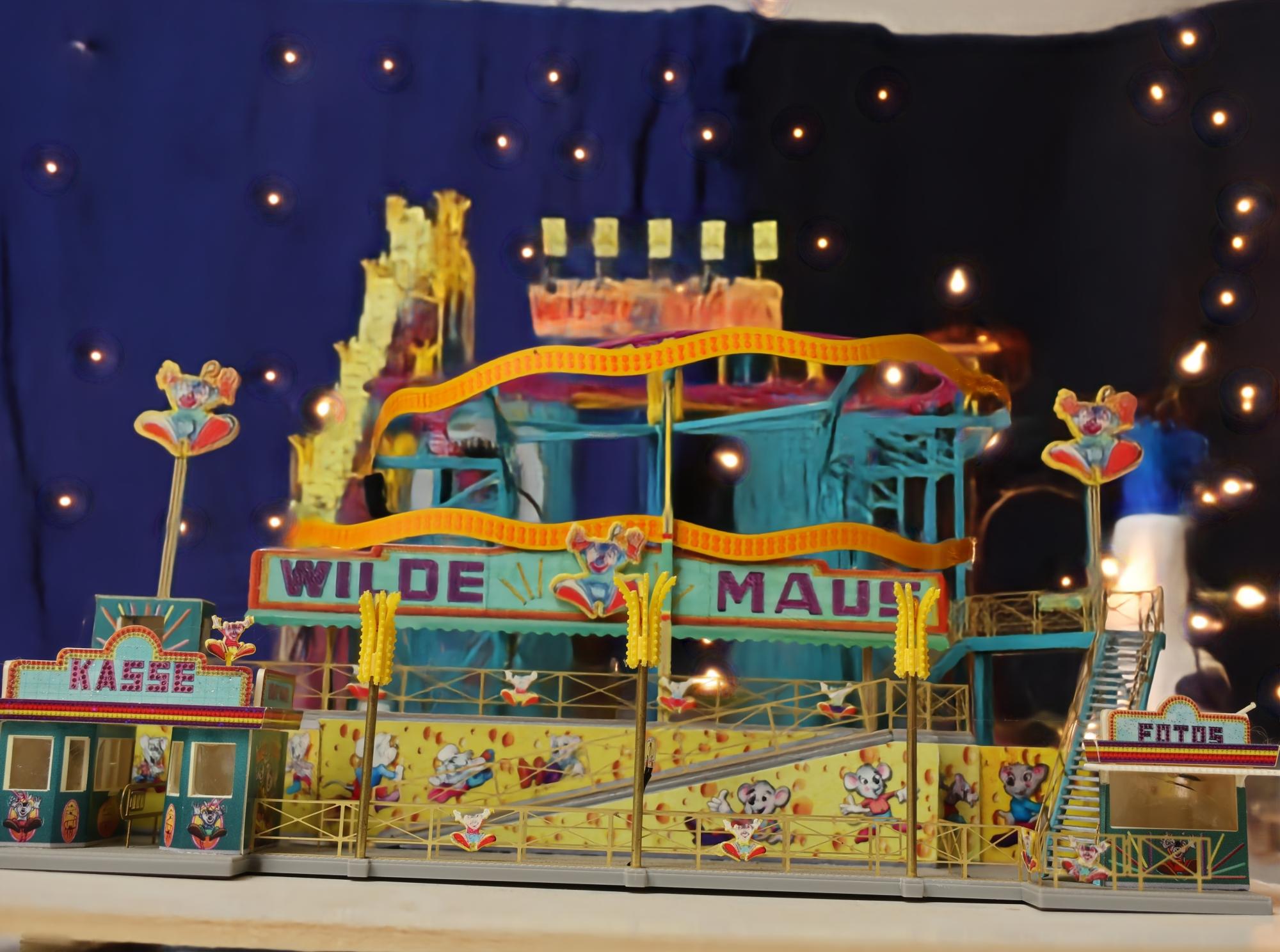} &
            \includegraphics[width=\widthcompp\textwidth,valign=t, trim={650px 940px 1000px 332px},clip]{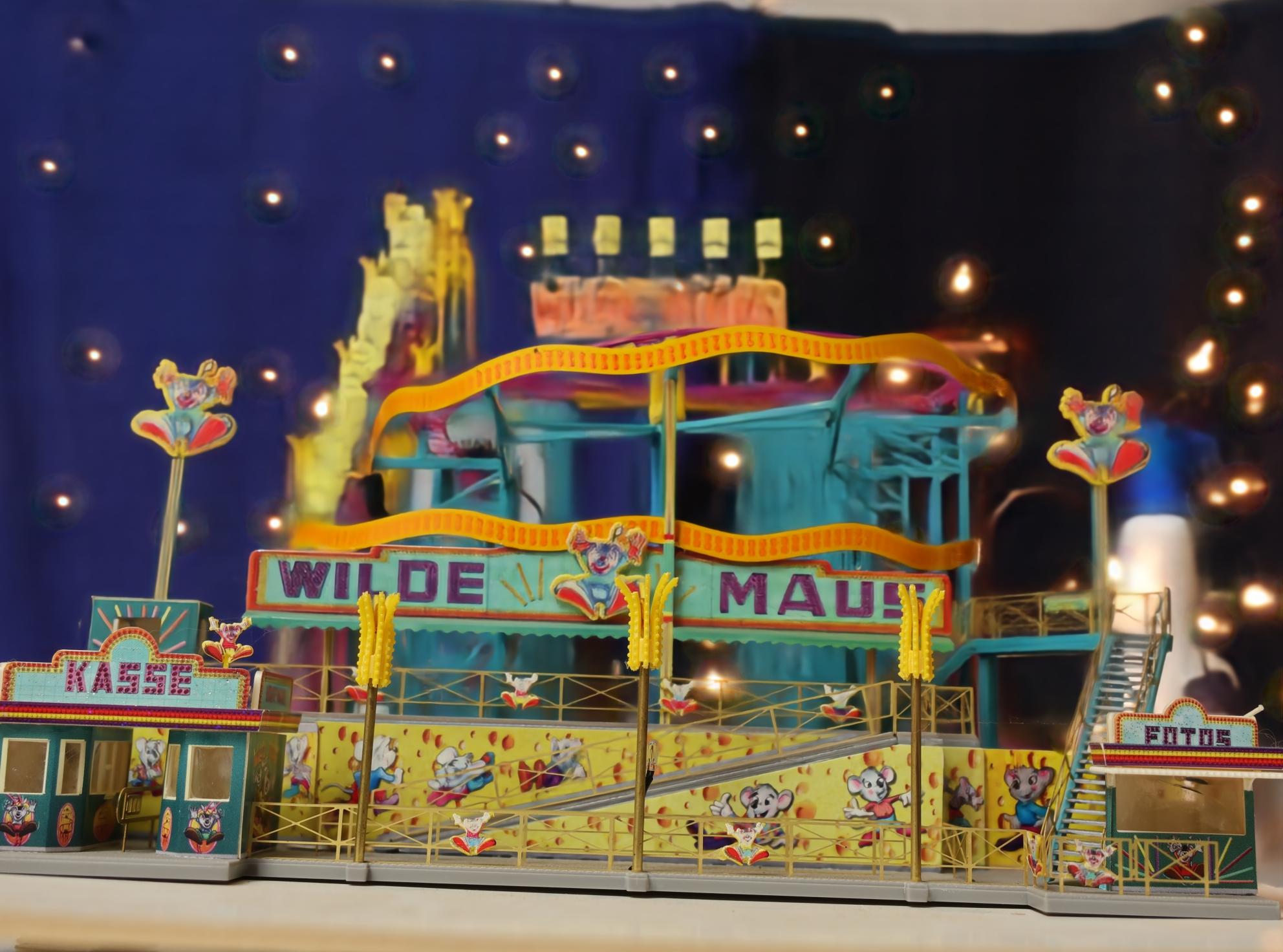}
            \\
            \addlinespace[2.0pt]
            &
            \includegraphics[width=\widthcompp\textwidth,valign=t, trim={650px 950px 1000px 340px},clip]{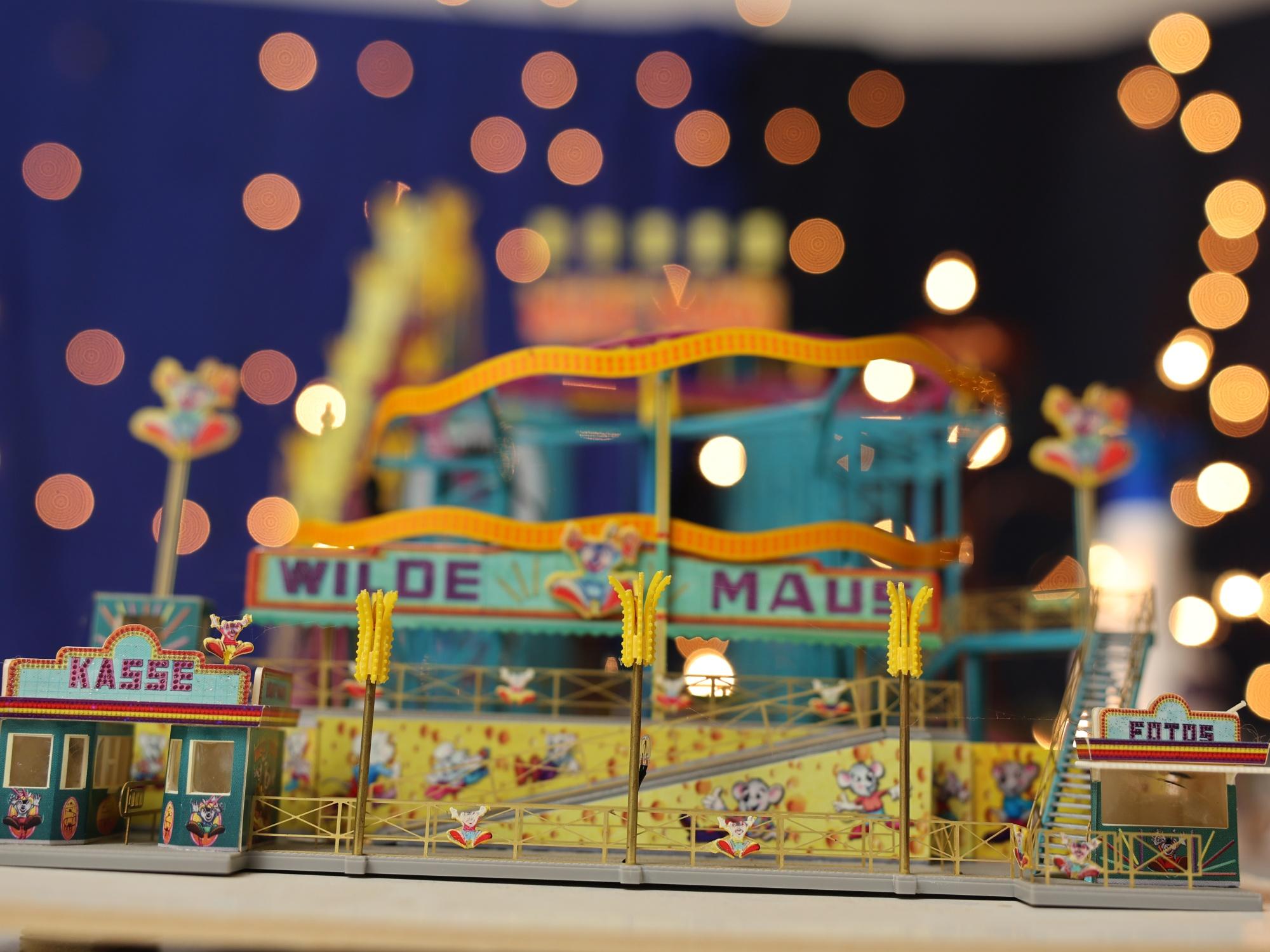} &
            \includegraphics[width=\widthcompp\textwidth,valign=t, trim={650px 944px 1000px 334px},clip]{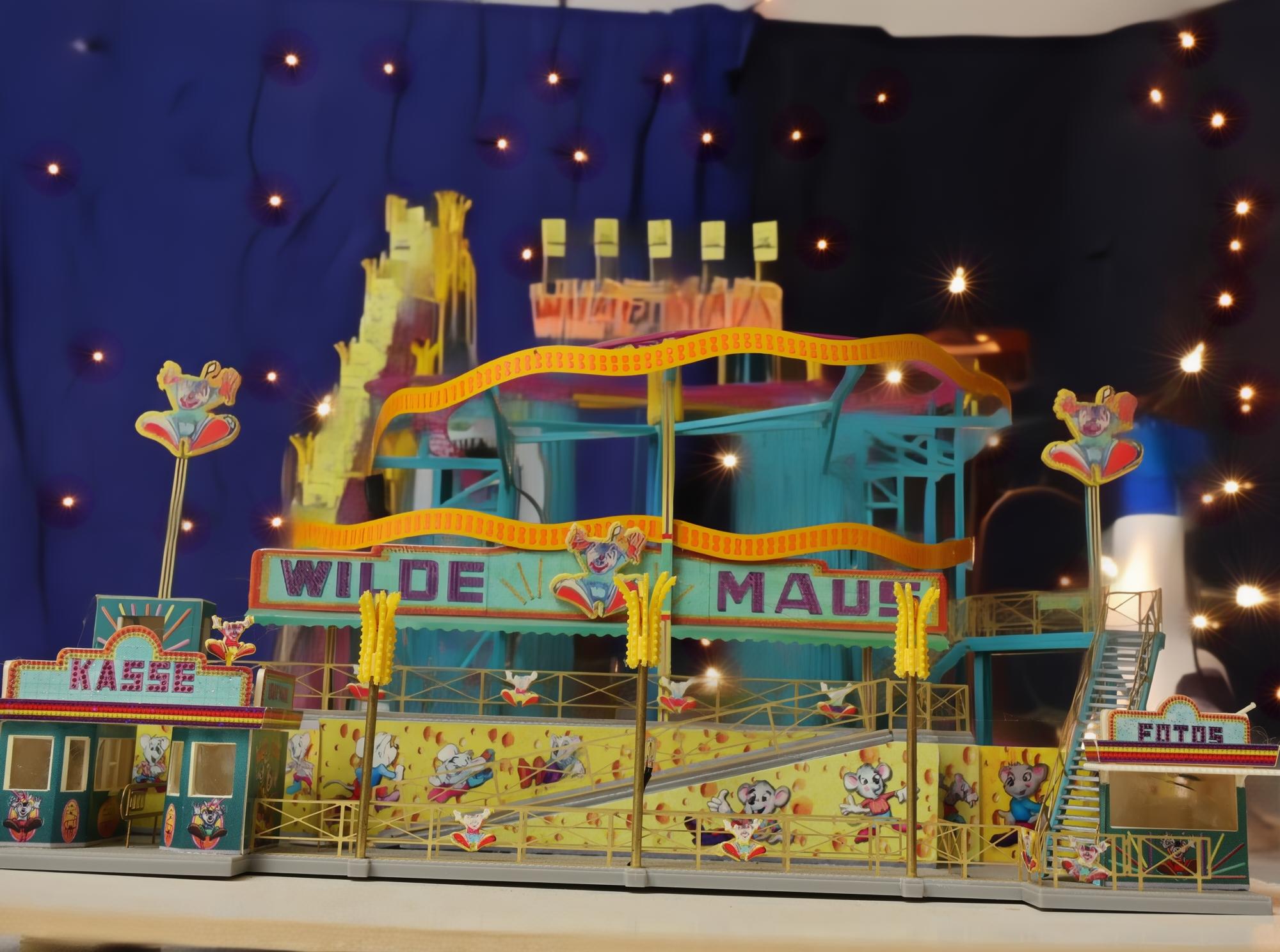} &
            \includegraphics[width=\widthcompp\textwidth,valign=t, trim={650px 944px 1000px 334px},clip]{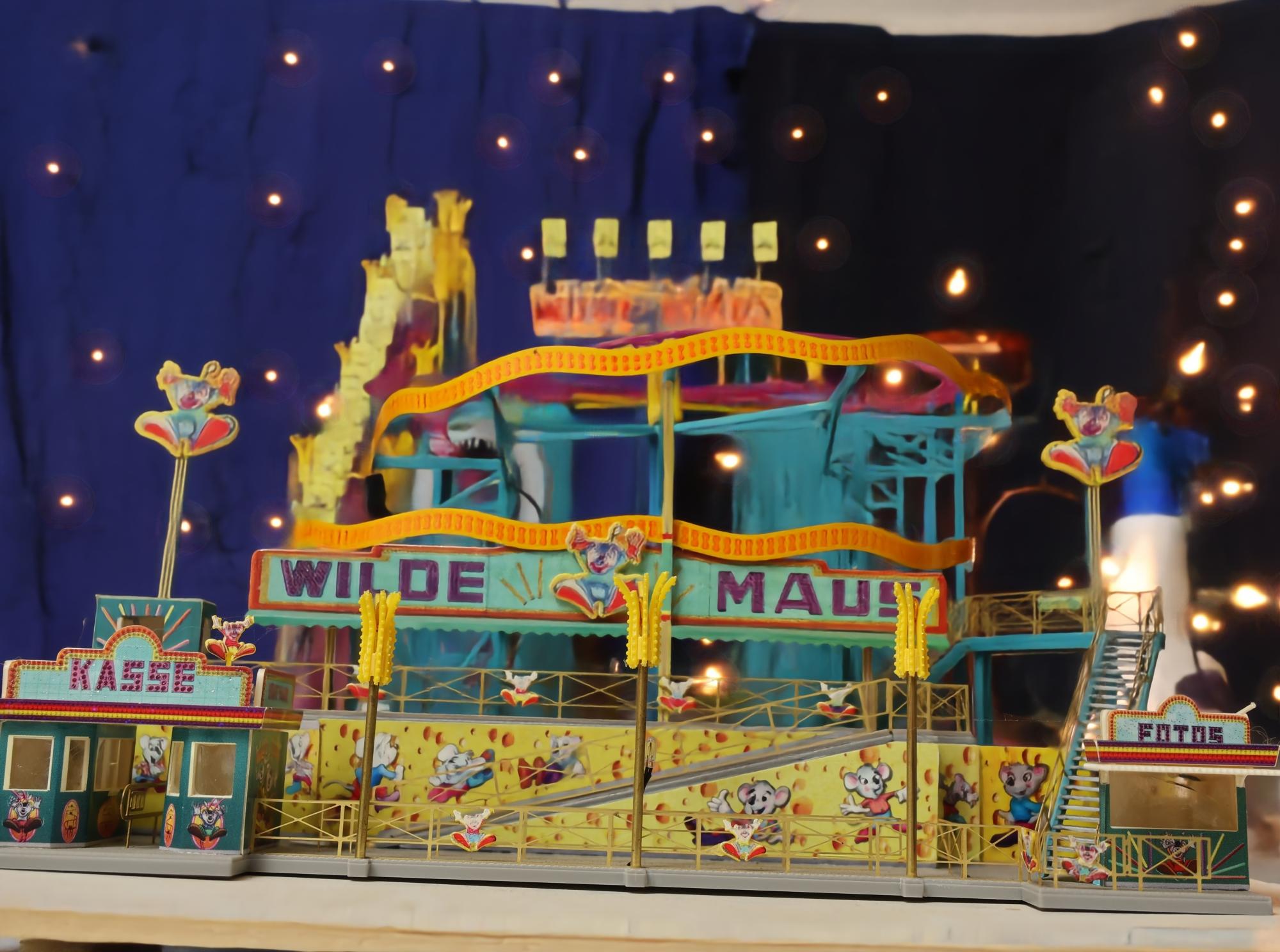} &
            \includegraphics[width=\widthcompp\textwidth,valign=t, trim={650px 950px 1000px 340px},clip]{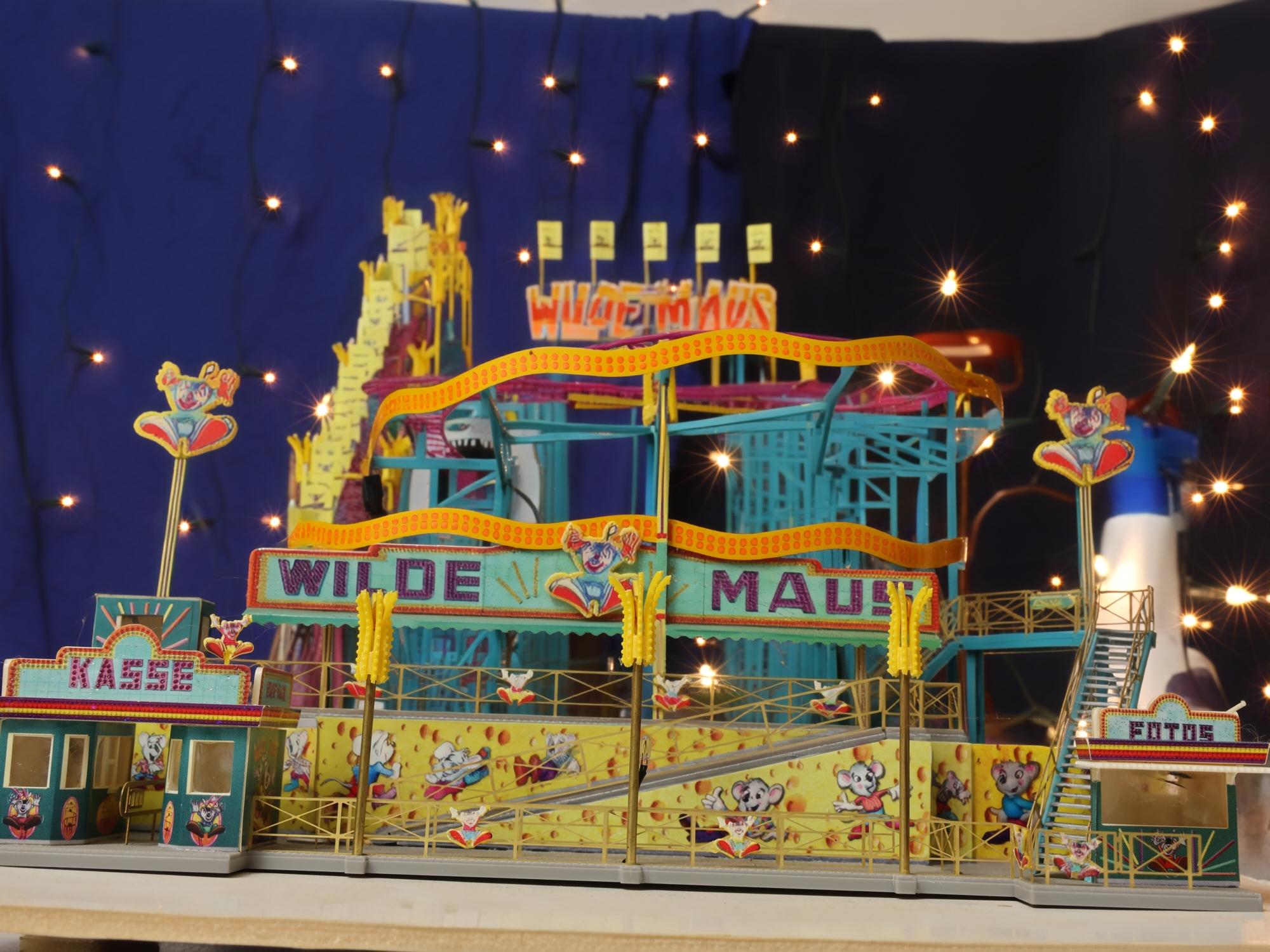} &
            \includegraphics[width=\widthcompp\textwidth,valign=t, trim={650px 940px 1000px 332px},clip]{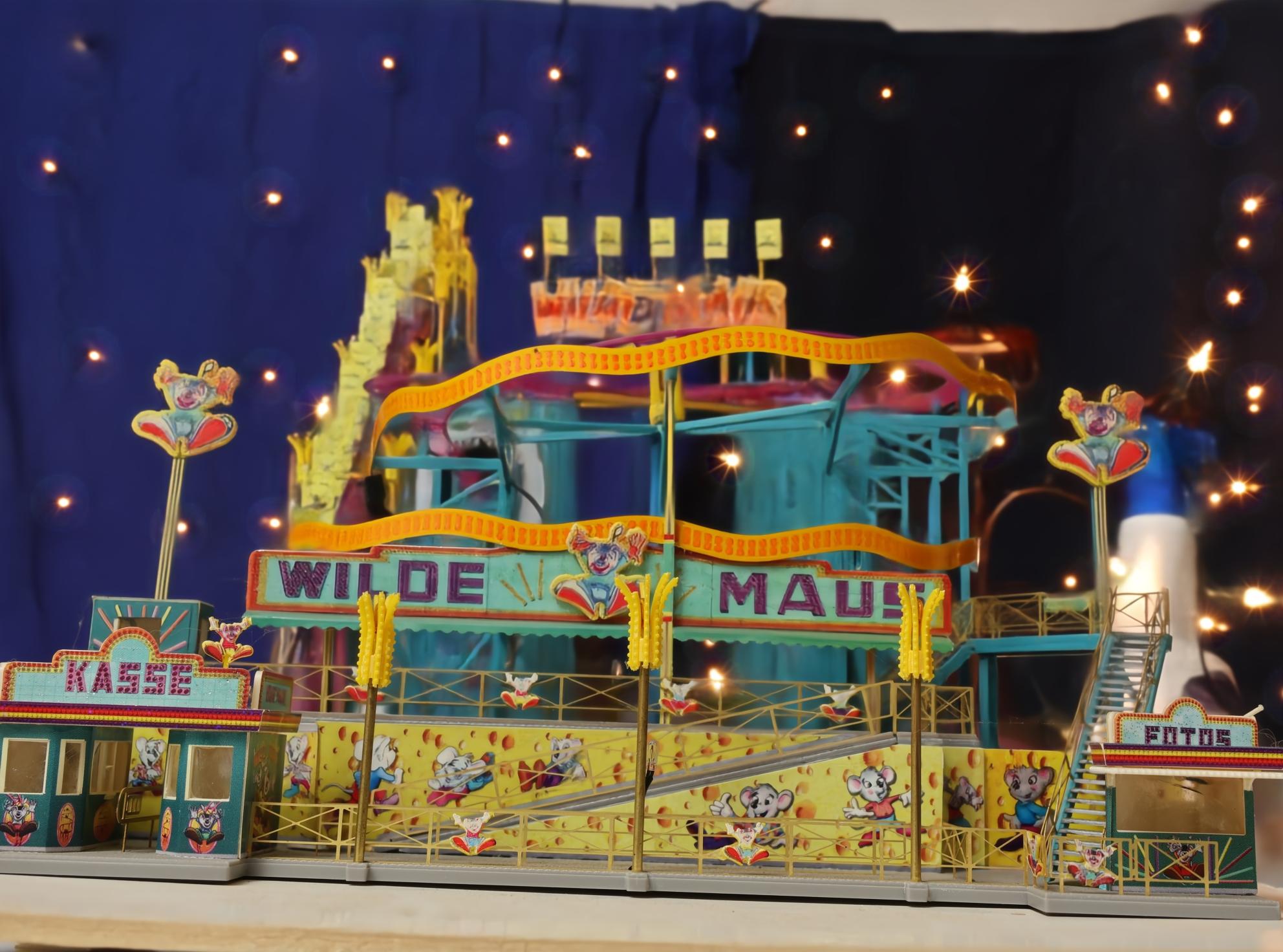}
            \\
            \addlinespace[1pt]
            & \fnum{2.8} Input: 16.34 & LAKDNet: 25.11 & EAMamba: 25.25 & Bokehlicious: 25.87 & FFTFormer: 26.06 \\

    \end{tabular}
\end{center}
\vspace{-6mm}
\caption{\textbf{Qualitative comparison on the RealDefocus~\cite{seizinger2025bokehlicious} test set for two aperture settings (\fnum{2.0} and \fnum{2.8}).} For each example, we report PSNR (dB) with respect to the sharp ground truth (\fnum{22.0}). The visual results highlight differences in detail recovery and texture fidelity across representative defocus-specific, general restoration, and rendering-based methods. Zooming in is recommended to better observe fine structures and reconstruction quality.}
\vspace{-1mm}
\label{fig:RealDefocusQual}
\end{figure*}

\subsection{Model Selection}

To provide a comprehensive and representative benchmark, we evaluate methods from four complementary categories: (i) dedicated Defocus Deblurring approaches, (ii) motion deblurring methods,  (iii) general-purpose image restoration architectures, and (iv) a Bokeh Rendering model. This selection reflects both historical developments in Defocus Deblurring and the recent trend toward large, versatile backbones that generalize across restoration tasks.

For (i) dedicated Defocus Deblurring approaches, we consider both early neural approaches and more recent physics-inspired designs.
Early Defocus Deblurring approaches were predominantly model-based and followed a two-stage pipeline: first estimating a spatially varying blur kernel for each pixel, and subsequently performing non-blind deconvolution~\cite{levin2007image}. Kernel estimation commonly relied on simplified parametric assumptions, such as Gaussian~\cite{karaali2017edge} or disk-shaped point spread functions~\cite{d2016non}, reflecting the aperture geometry of camera lenses. While principled, these approaches were limited by the accuracy of kernel estimation and their sensitivity to noise and model mismatch.

With the advent of deep learning, end-to-end CNN-based methods replaced explicit kernel estimation by directly learning the mapping from blurry to sharp images. Early examples include AIFNet~\cite{ruan2021aifnet} and DPDNet~\cite{abuolaim2020defocus}. DPDNet, in particular, leveraged special dual-pixel (DP) sub-aperture views to better constrain the ill-posed inversion, representing one of the first neural frameworks specifically designed for Defocus Deblurring.
Subsequent works reintroduced explicit blur modeling within deep architectures:
GKMNet~\cite{quan2021gaussian} explicitly models spatially variant defocus blur kernels using a compact linear parametric form and employs a lightweight scale-recurrent architecture with multi-scale supervision; NRKNet~\cite{quan2023neumann} further incorporates a learnable recursive kernel representation inspired by truncated Neumann series for approximating blur operator inversion, combined with cross-scale fusion and a reblurring loss for regularization.
Another line of work focuses on data-driven architectural design rather than explicit kernel formulation: DRBNet~\cite{ruan2022learning} adopts dynamic filtering mechanisms~\cite{lee2021iterative} tailored to defocus characteristics, while LAKDNet~\cite{ruan2023revisiting} employs large-kernel depth-wise convolutions and spatial-channel mixing to emulate the large effective receptive field of transformers~\cite{vaswani2017attention} within a purely convolutional framework. These approaches~\cite{ruan2022learning, ruan2023revisiting} are typically pretrained on synthetic data and subsequently fine-tuned on real-world datasets to mitigate domain gaps.

For (ii) motion deblurring methods, we include recent state-of-the-art architectures that have demonstrated strong generalization capability.
EVSSM~\cite{kong2025efficient} enhances State Space Model~\cite{gu2022efficiently} (SSM) based restoration by introducing geometric transformations prior to state-space modeling to better capture local structures, complemented by an efficient frequency-domain feedforward network for improved discriminative capacity.
FFTFormer~\cite{kong2023efficient} integrates frequency-domain processing within a transformer framework, enabling effective modeling of global blur patterns while maintaining high reconstruction fidelity.

For (iii) general-purpose restoration architectures, we include Restormer~\cite{zamir2022restormer} and EAMamba~\cite{lin2025eamamba}. Restormer is a transformer-based U-Net architecture that is widely used across restoration tasks, including motion and Defocus Deblurring, denoising, and deraining.
EAMamba builds on SSMs and introduces an enhanced vision Mamba~\cite{mamba} architecture tailored for image restoration. By leveraging long-range dependency modeling, it has shown strong performance across motion deblurring, dehazing, and super-resolution.

For (iv) Bokeh Rendering, we include Bokehlicious~\cite{seizinger2025bokehlicious}, which is built on a physics-informed transformer architecture that mimics the behavior of the aperture mechanism in real camera lenses.

\subsection{Training Protocol}

All methods are trained following their respective original implementations as closely as possible, particularly with respect to optimizer choice, learning rate schedules, and architectural details. Training is performed on RealDefocus at 2000$\times$1500 resolution with $512\times512$ patches and includes flips and rotations for data augmentation. Importantly, each method is optimized using the loss function specified in its original implementation. This is also necessary for approaches that rely on specialized objectives, such as multi-scale supervision in GKMNet~\cite{quan2021gaussian} and NRKNet~\cite{quan2023neumann}, or reblurring-based regularization in NRKNet~\cite{quan2023neumann}. 

\section{BENCHMARKING AND RESULTS ANALYSIS}
\label{sec:result}

Overall, the benchmark aims to ensure fairness while respecting the design principles of each method. By combining physics-driven defocus models, motion-deblurring state-of-the-art architectures, general-purpose restoration backbones, and a Bokeh Rendering model, we provide a broad and balanced evaluation of current approaches under a unified experimental setting.

\begin{table*}[t]
        \centering
        \small
        \begin{tabular}{l|ccc|ccc}
            \toprule
            \multirow{2}{*}{Method}                 & \multicolumn{3}{c|}{Trained on DPDD$_S$~\cite{abuolaim2020defocus}}              & \multicolumn{3}{c}{Trained on RealDefocus~\cite{seizinger2025bokehlicious}}                                                               \\
                                                    & PSNR$\uparrow$    & SSIM$\uparrow$    & LPIPS$\downarrow$ & PSNR$\uparrow$ (change)           & SSIM$\uparrow$ (change)                   & LPIPS$\downarrow$ (change)        \\
            \midrule
           %XXXNet~\cite{}                          & 00.000            & 0.000             & 0.000             & 00.000 (\textcolor{teal}{$+$0.000})   & 0.000 (\textcolor{teal}{$+$0.000})    & 0.000 (\textcolor{teal}{$-$0.000})  \\
            % \rowcolor{lgray}
            DPDNet$_S$~\cite{abuolaim2020defocus}   & 22.870            & 0.670             & 0.425             & 23.934 (\textcolor{teal}{$+$1.064})   & 0.714 (\textcolor{teal}{$+$0.044})    & 0.418 (\textcolor{teal}{$-$0.007})  \\
            \rowcolor{lgray}
            GKMNet~\cite{quan2021gaussian}          & 24.254            & 0.732             & 0.392             & 25.014 (\textcolor{teal}{$+$0.760})   & 0.755 (\textcolor{teal}{$+$0.023})    & 0.335 (\textcolor{teal}{$-$0.057})  \\
            DRBNet~\cite{ruan2022learning}          & 24.884            & 0.751             & 0.376             & 25.283 (\textcolor{teal}{$+$0.399})   & 0.774 (\textcolor{teal}{$+$0.023})    & 0.272 (\textcolor{teal}{$-$0.104})  \\
            \rowcolor{lgray}
            Restormer~\cite{zamir2022restormer}     & 25.080            & 0.769             & 0.289             & 25.382 (\textcolor{teal}{$+$0.302})   & 0.797 (\textcolor{teal}{$+$0.028})    & 0.268 (\textcolor{teal}{$-$0.021})  \\
            NRKNet~\cite{quan2023neumann}           & 25.148            & 0.768             & 0.338             & 25.397 (\textcolor{teal}{$+$0.249})   & 0.784 (\textcolor{teal}{$+$0.016})    & 0.313 (\textcolor{teal}{$-$0.025})  \\
            \rowcolor{lgray}
            LAKDNet~\cite{ruan2023revisiting}       & 25.080            & 0.762             & 0.267             & 25.413 (\textcolor{teal}{$+$0.333})   & 0.787 (\textcolor{teal}{$+$0.025})    & \underline{0.242} (\textcolor{teal}{$-$0.025})  \\
            EAMamba~\cite{lin2025eamamba}           & 24.685            & 0.755             & 0.354             & 25.656 (\textcolor{teal}{$+$0.971})   & \underline{0.801} (\textcolor{teal}{$+$0.046})& 0.263 (\textcolor{teal}{$-$0.091})\\
            \rowcolor{lgray}
            FFTFormer~\cite{kong2023efficient}      & 24.938            & 0.764             & 0.353             &\textit{25.710} (\textcolor{teal}{$+$0.772})&\textbf{0.804} (\textcolor{teal}{$+$0.040})&\textit{0.252} (\textcolor{teal}{$-$0.101})\\
            EVSSM~\cite{kong2025efficient}          & 25.087            & 0.765             & 0.340             & \underline{25.798} (\textcolor{teal}{$+$0.711})           & \textit{0.800} (\textcolor{teal}{$+$0.035})& 0.279 (\textcolor{teal}{$-$0.061})          \\
            \rowcolor{lgray}
            Bokehlicious~\cite{seizinger2025bokehlicious}& 24.456       & 0.754             & 0.249             & \textbf{25.858} (\textcolor{teal}{$+$1.402}) & 0.797 (\textcolor{teal}{$+$0.043})& \textbf{0.205} (\textcolor{teal}{$-$0.044})\\
            \bottomrule
        \end{tabular}
        \vspace{-2mm}
        \caption{
        \textbf{Quantitative results on the RealDOF dataset~\cite{lee2021iterative} when trained on DPDD$_S$~\cite{abuolaim2020defocus} and on RealDefocus~\cite{seizinger2025bokehlicious}.} We additionally report the performance change with respect to models trained on DPDD$_S$. Some DPDD$_S$-trained RealDOF results are taken from~\cite{quan2023neumann, ruan2023revisiting, quan2023single}. The best, second-best, and third-best results are indicated in \textbf{bold}, \underline{underline}, and \textit{italic}, respectively.
        }
        \label{tab:RealDOF}
        \vspace{-5mm}
\end{table*}

\begin{figure*}[t]
\vspace{3mm}
\begin{center}
\renewcommand{\arraystretch}{0.4}
\setlength{\tabcolsep}{0.8pt}
\footnotesize
\def\imgcompp{.1752}
\def\widthcompp{.14}
\begin{tabular}[b]{c@{ } c@{ }  c@{ } c@{ } c@{ }  c@{ }c}\hspace{-4mm}
Input &  GT: PSNR (dB) & EAMamba: 22.45 & FFTFormer: 24.41 & EVSSM: 24.74 & Bokehlicious: 23.67 \\
\addlinespace[0.75pt]
\multirow{3}{*}[1.5mm]{\includegraphics[height=\imgcompp\textwidth, trim={200px 100px 600px 200px},clip]{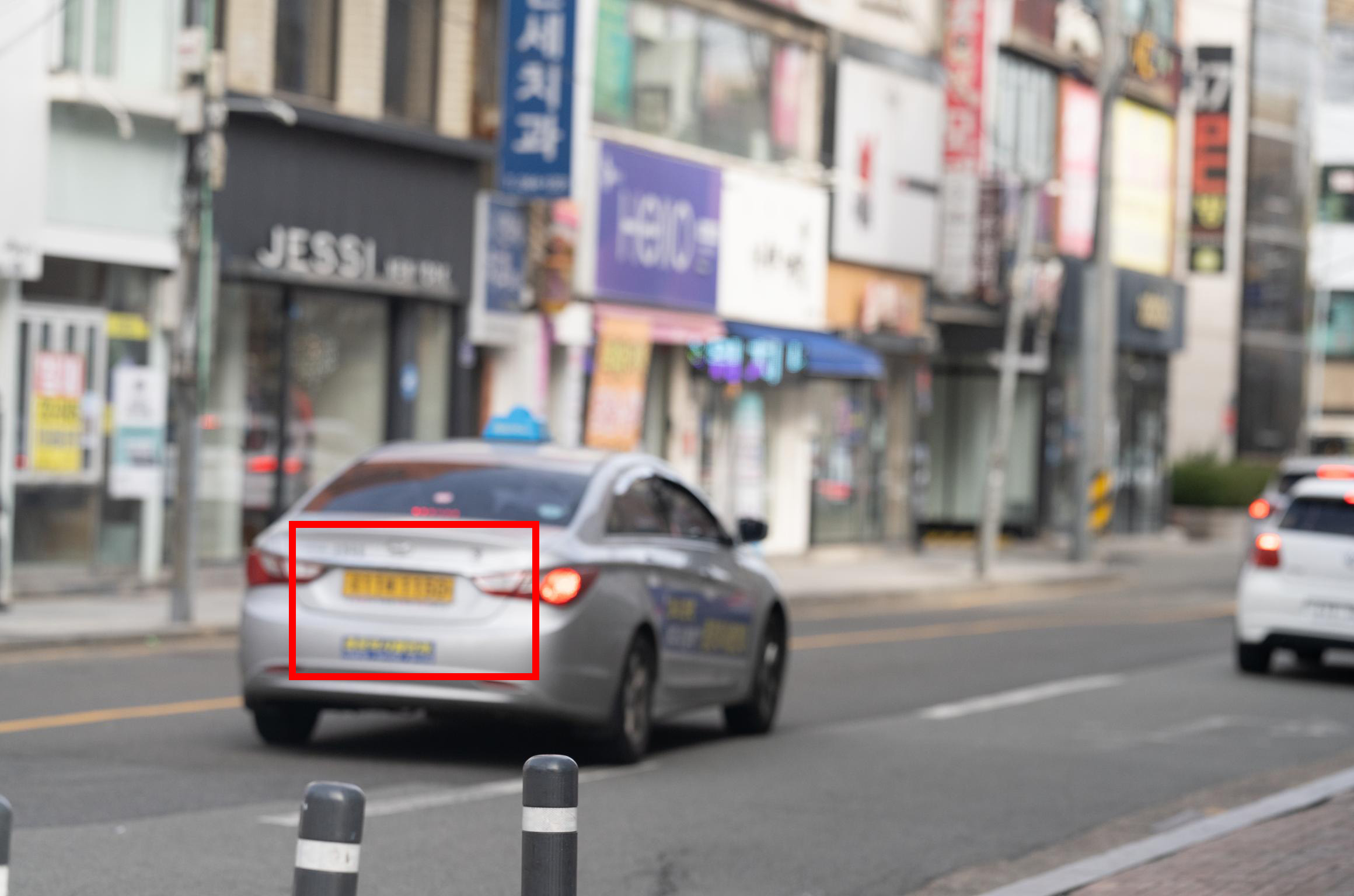}} &
            \includegraphics[width=\widthcompp\textwidth,valign=t, trim={500px 380px 1400px 900px},clip]{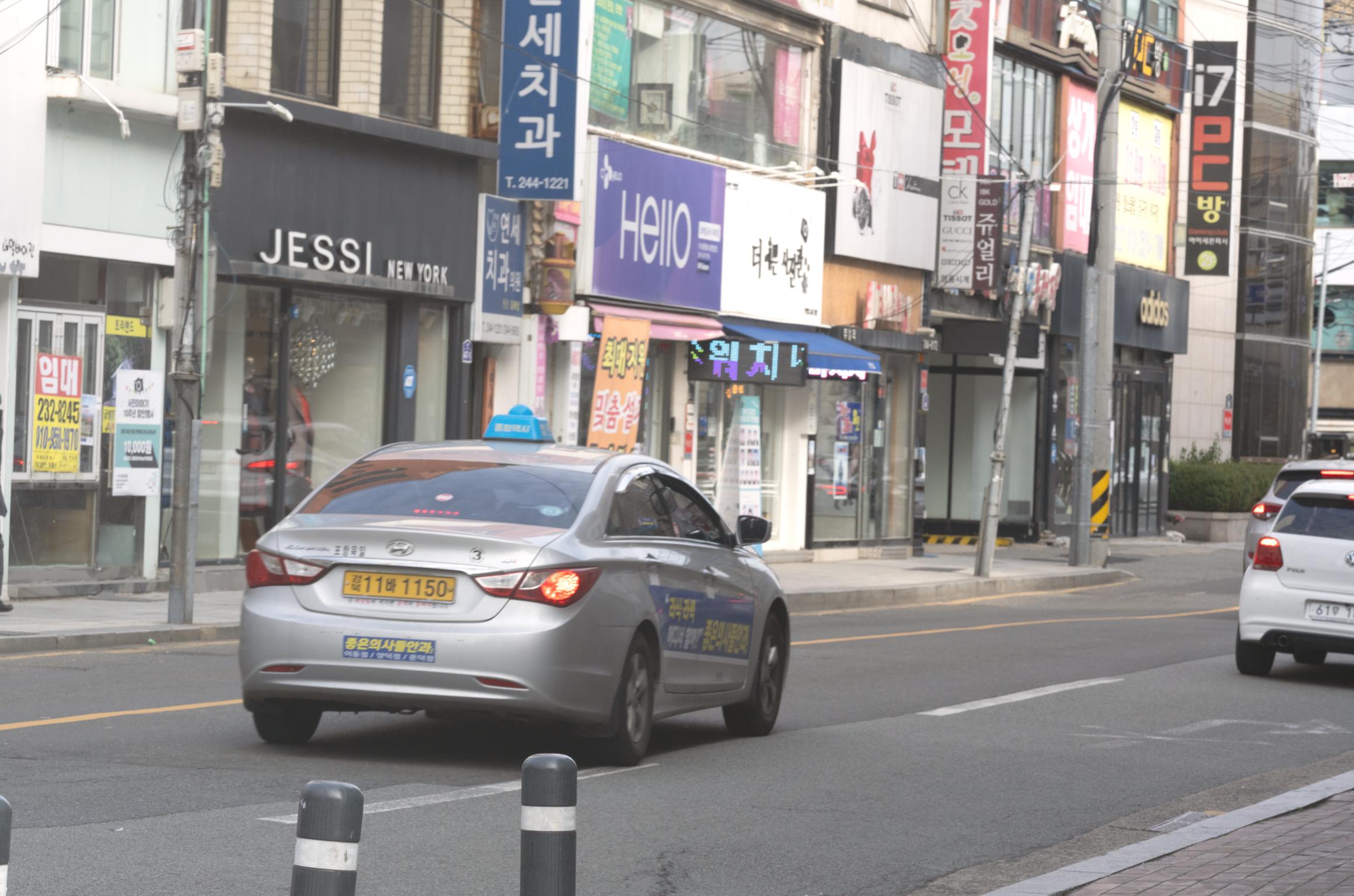} &
            \includegraphics[width=\widthcompp\textwidth,valign=t, trim={500px 380px 1400px 900px},clip]{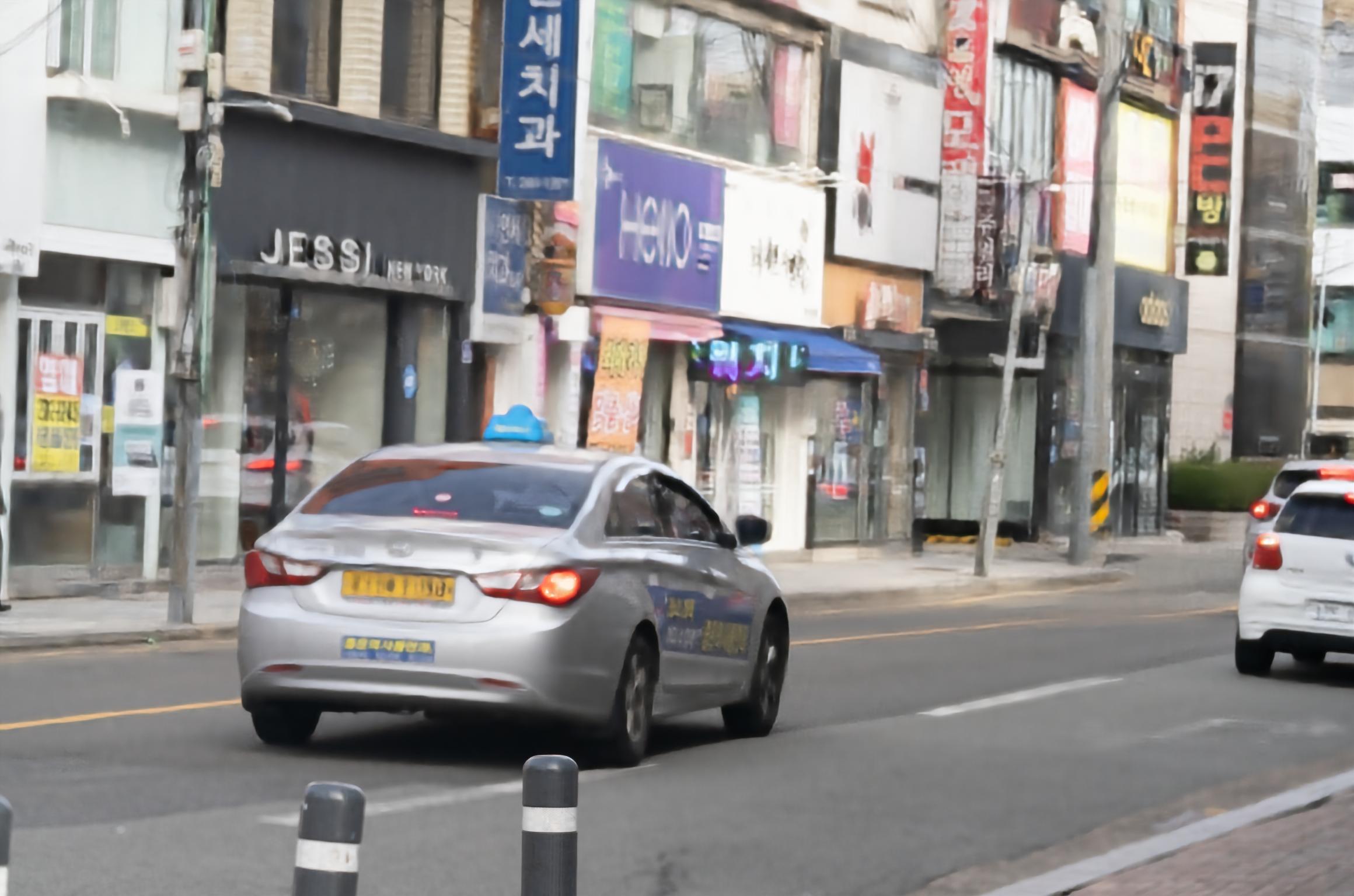} &
            \includegraphics[width=\widthcompp\textwidth,valign=t, trim={492px 380px 1392px 900px},clip]{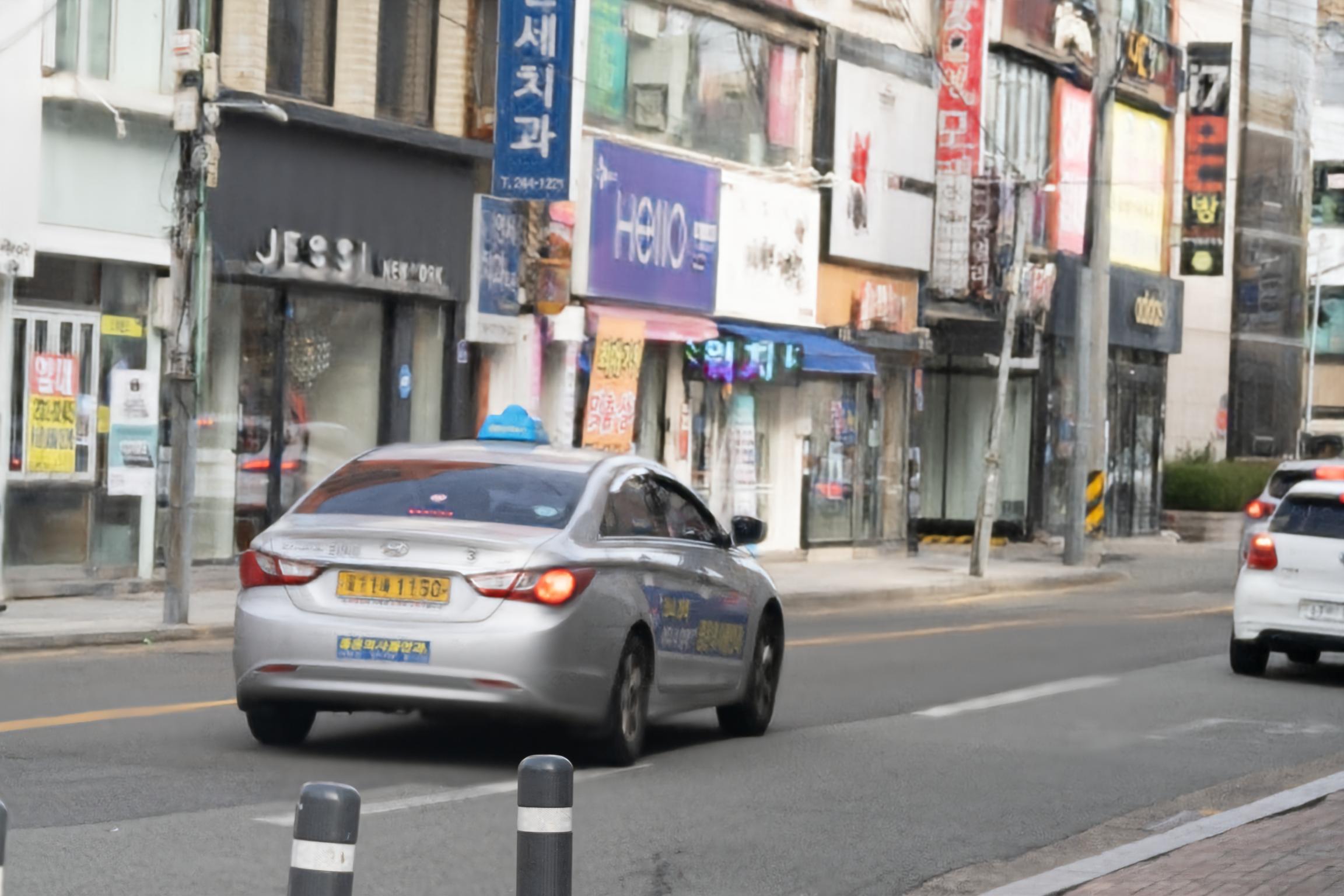} &
            \includegraphics[width=\widthcompp\textwidth,valign=t, trim={492px 380px 1392px 900px},clip]{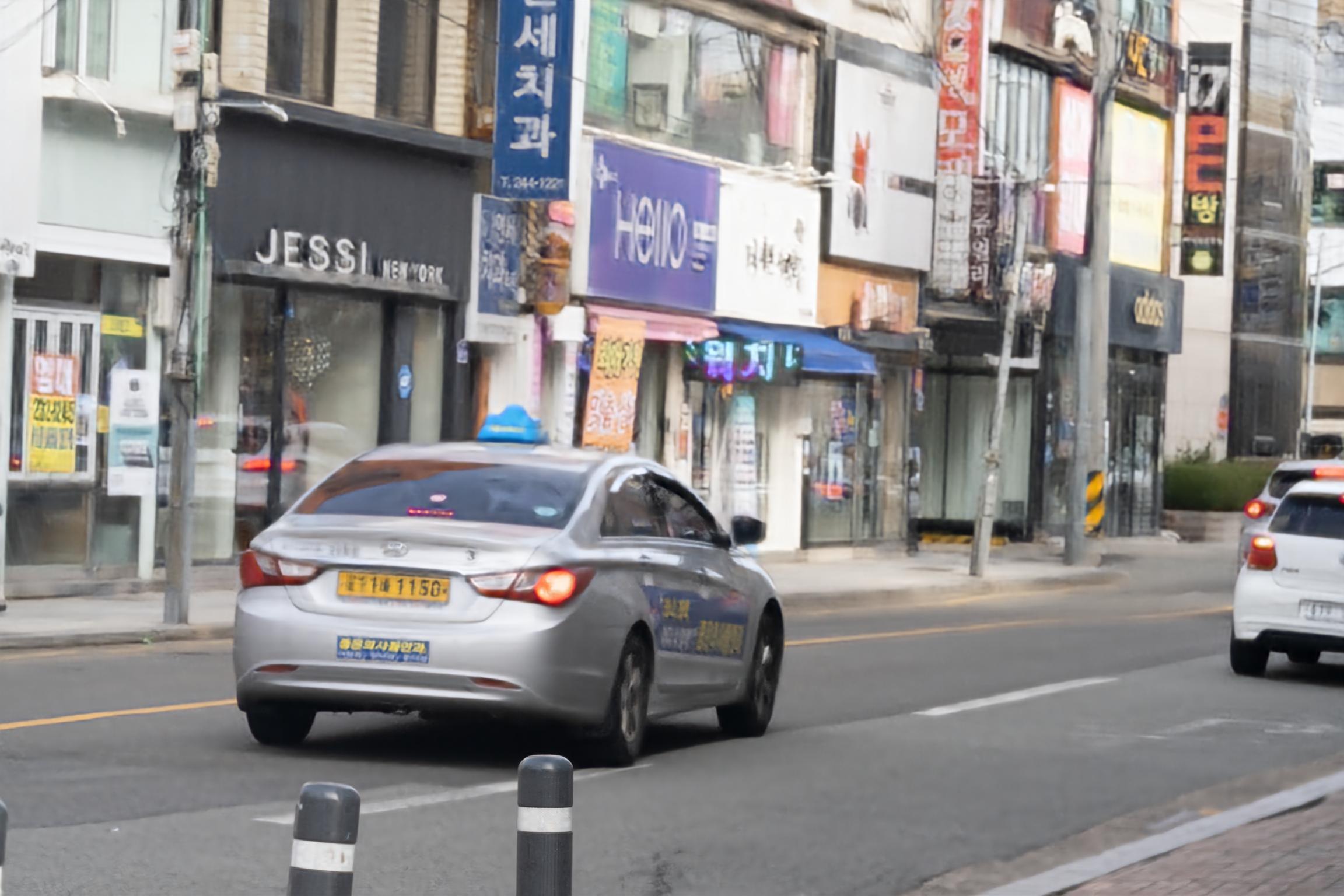} &
            \includegraphics[width=\widthcompp\textwidth,valign=t, trim={500px 380px 1400px 900px},clip]{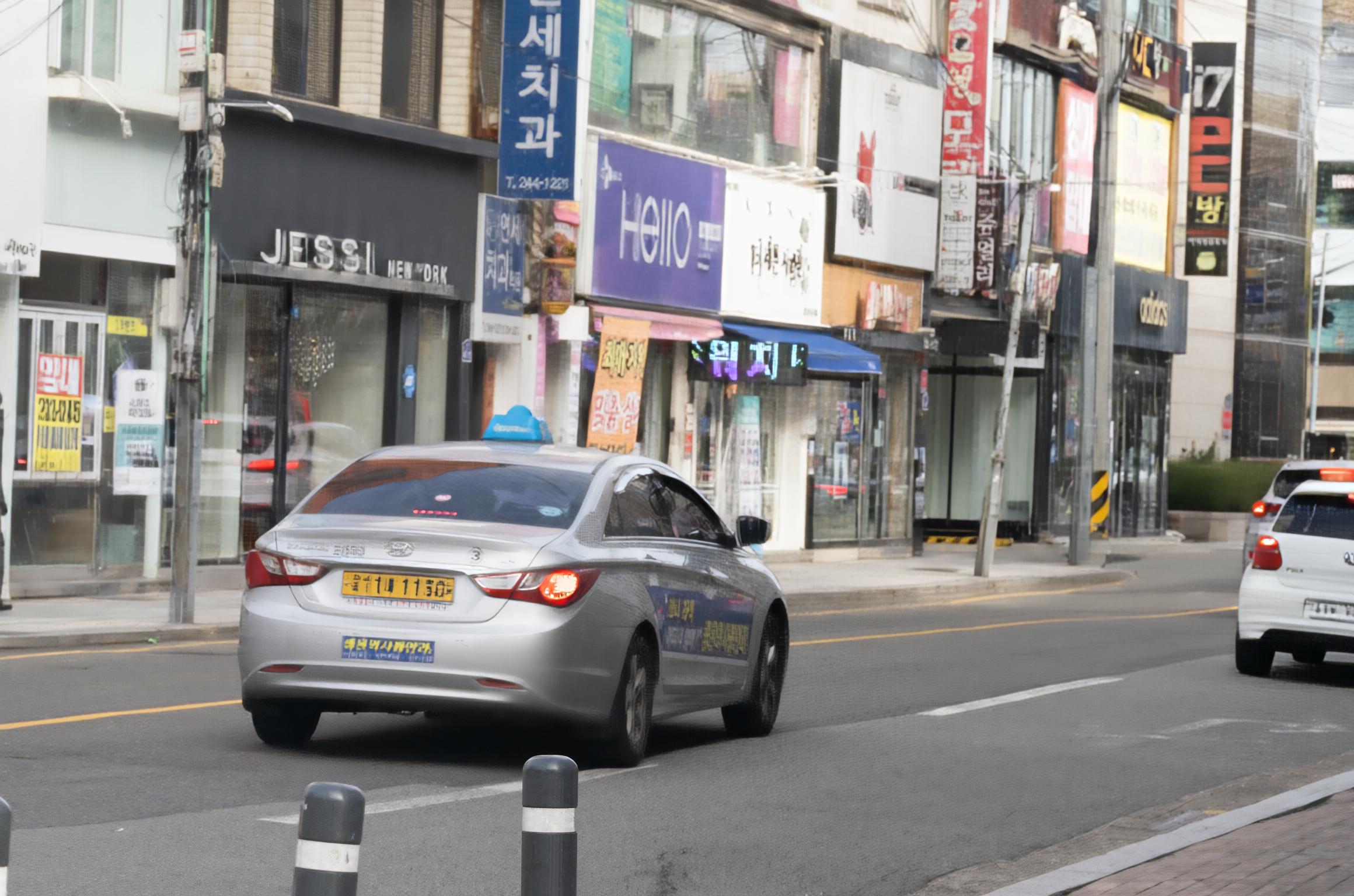} &
            \rotatebox{-90}{\hspace{1.5mm}\dpdd}
            \\
            \addlinespace[2.0pt]
            &
            \includegraphics[width=\widthcompp\textwidth,valign=t, trim={500px 380px 1400px 900px},clip]{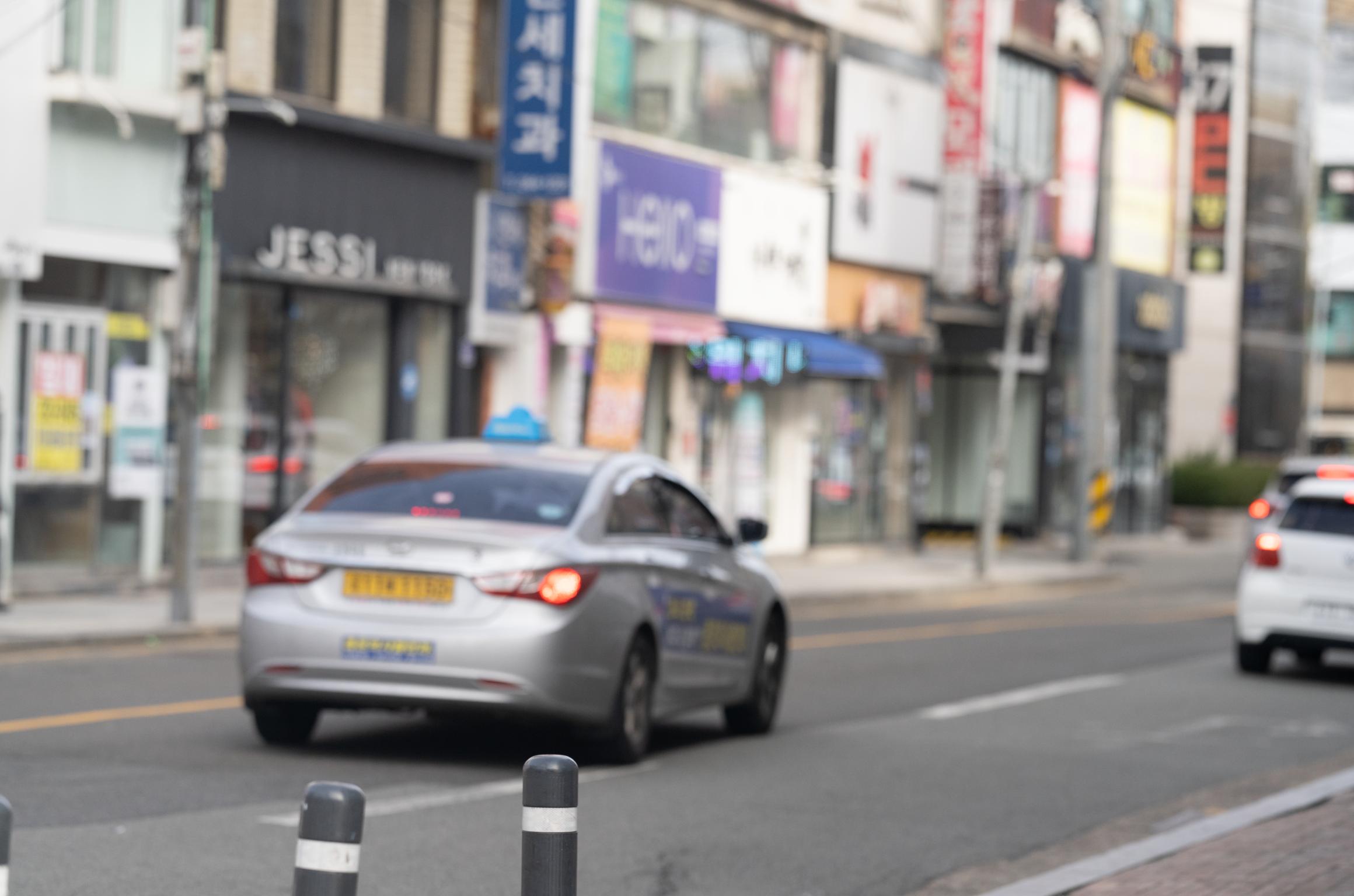} &
            \includegraphics[width=\widthcompp\textwidth,valign=t, trim={500px 380px 1400px 900px},clip]{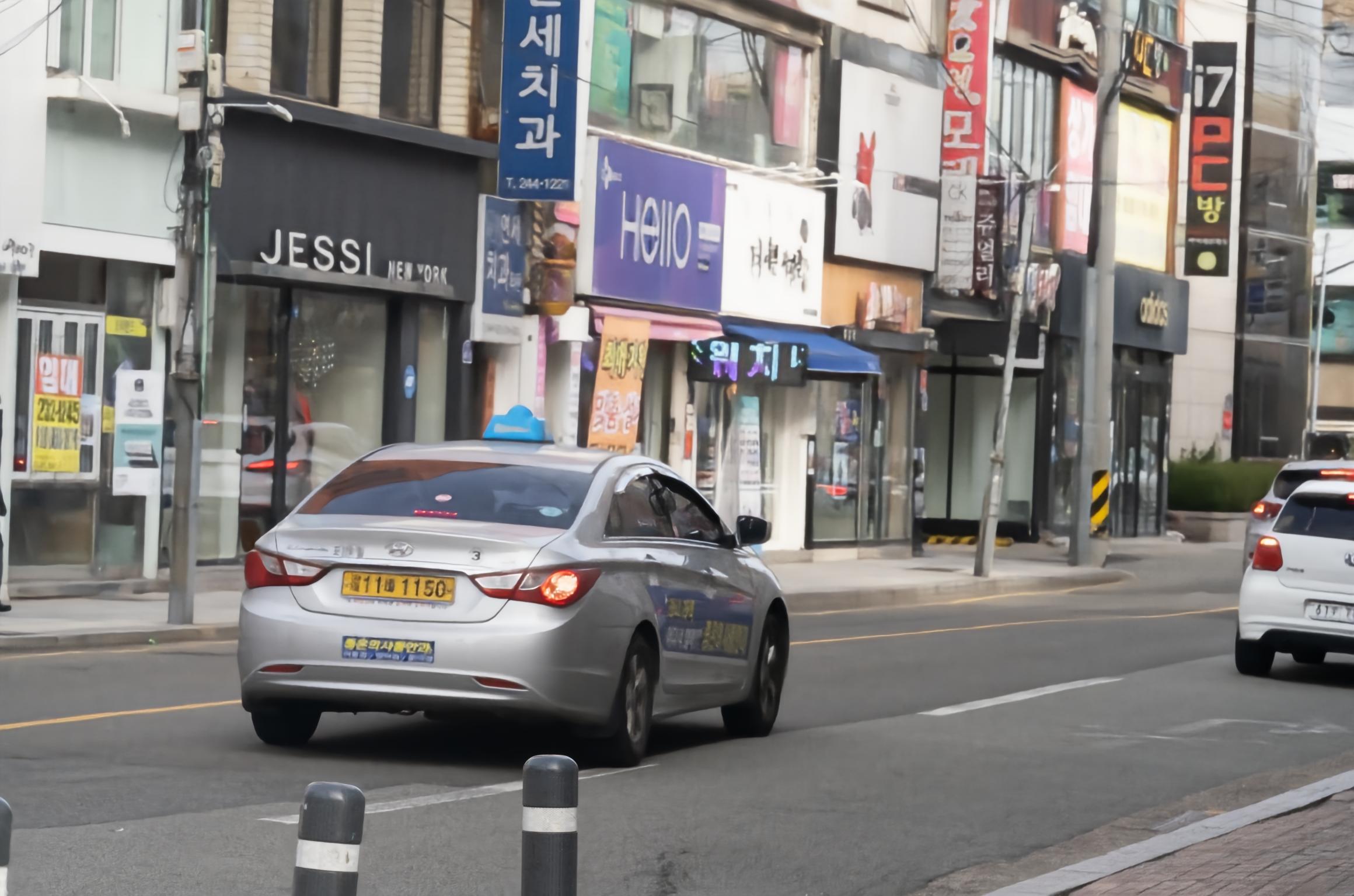} &
            \includegraphics[width=\widthcompp\textwidth,valign=t, trim={492px 380px 1392px 900px},clip]{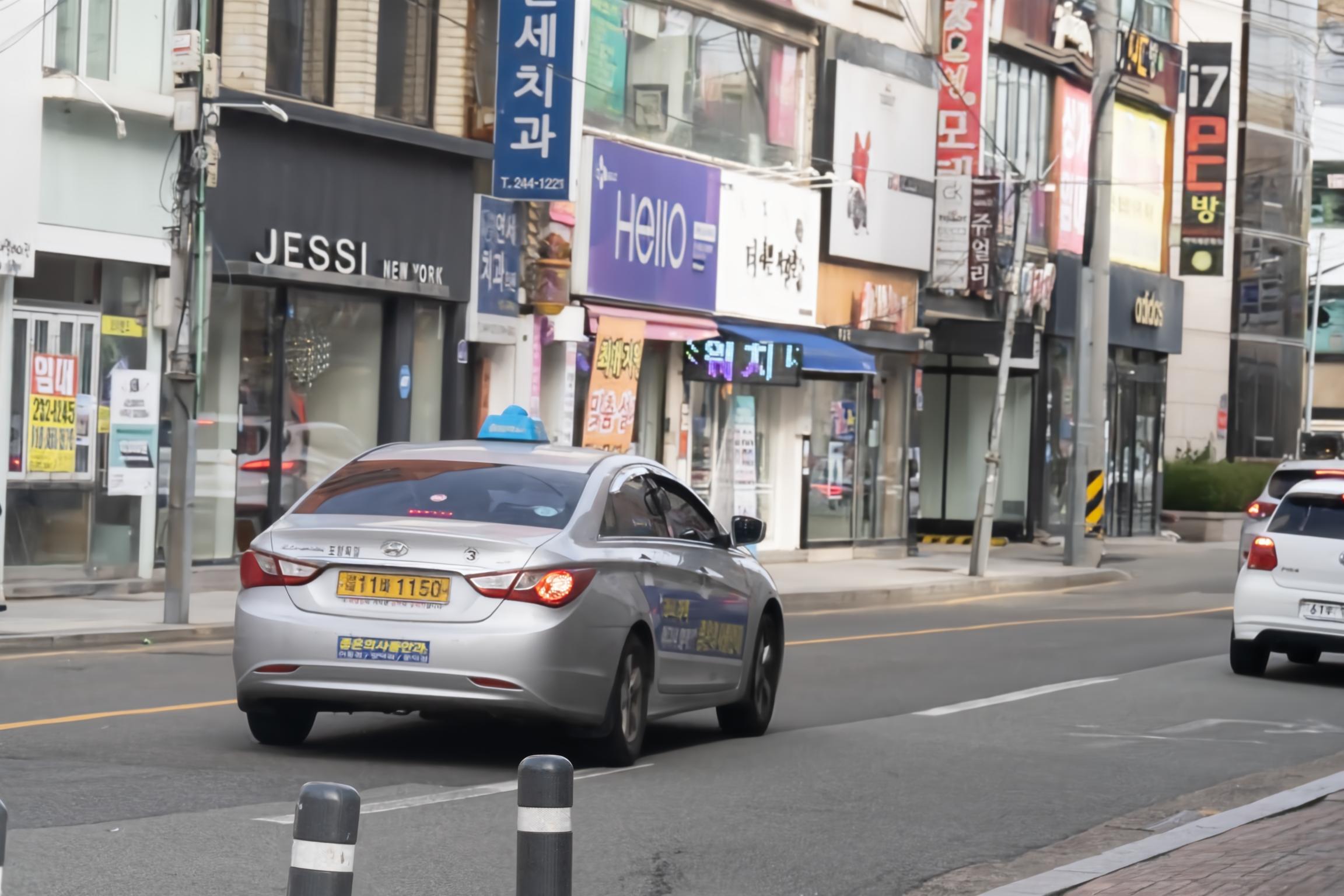} &
            \includegraphics[width=\widthcompp\textwidth,valign=t, trim={492px 380px 1392px 900px},clip]{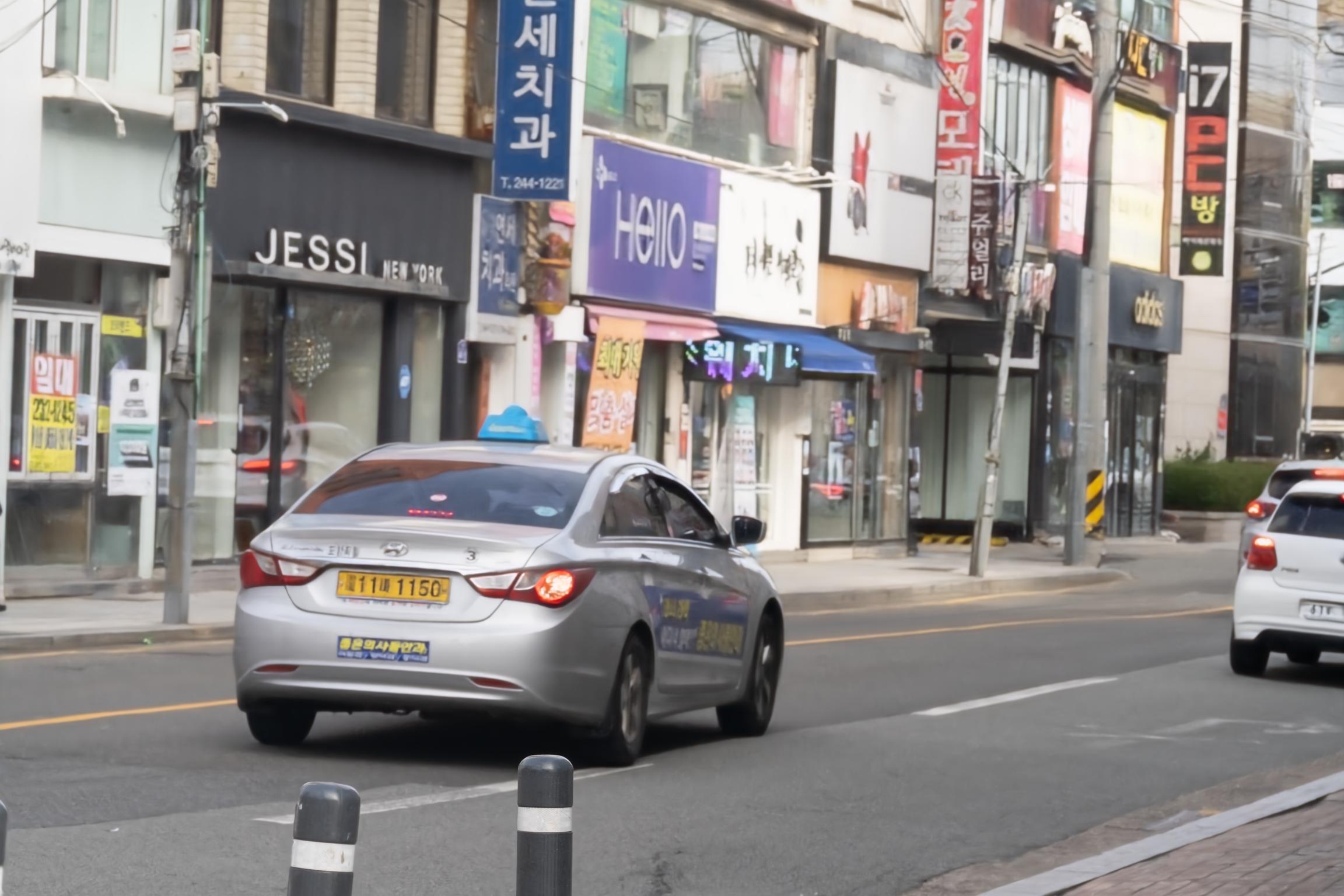} &
            \includegraphics[width=\widthcompp\textwidth,valign=t, trim={500px 380px 1400px 900px},clip]{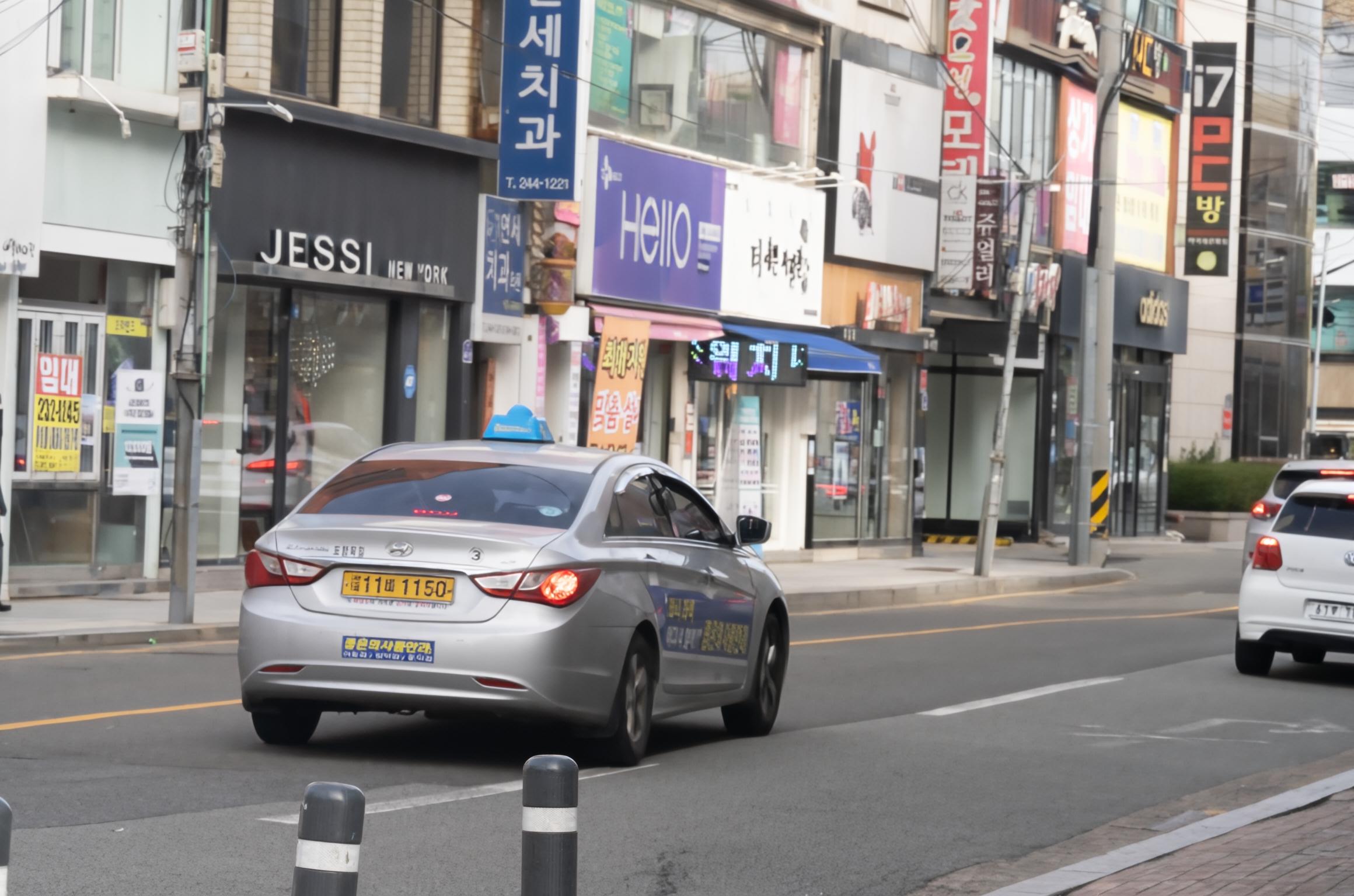} &
            \rotatebox{-90}{\hspace{-2.2mm}RealDefocus}
            \\
            \addlinespace[1pt]
            & Input: 21.57 & EAMamba: 24.45 & FFTFormer: 25.15 & EVSSM: 25.54 & Bokehlicious: 24.62 \\
\addlinespace[3.0pt]

    \end{tabular}
\end{center}
\vspace{-8mm}
\caption{\textbf{Qualitative comparison on the RealDOF~\cite{lee2021iterative} dataset.} For each example, the top row shows results of models trained on \dpdd~\cite{abuolaim2020defocus}, while the bottom row presents the same architectures trained on RealDefocus~\cite{seizinger2025bokehlicious}. Training on RealDefocus consistently improves reconstruction quality, yielding sharper details, as reflected by higher PSNR scores across methods. Zooming in is recommended to inspect fine structures and texture restoration.}
\vspace{-3.5mm}
\label{fig:RealDOFQual}
\end{figure*}

\subsection{The RealDefocus Benchmark}

Table~\ref{tab:RealDefocusBench} summarizes the performance of recent SIDD methods on the RealDefocus~\cite{seizinger2025bokehlicious} benchmark across different aperture settings. As expected, smaller \fstop~values (e.g., \fnum{2.0}) correspond to substantially stronger blur and thus represent the most challenging regime, while larger \fstops (e.g., \fnum{8.0}) are comparatively easier.

Averaged over all \fstops, FFTFormer~\cite{kong2023efficient} achieves the highest PSNR (30.206) and SSIM (0.8837), whereas Bokehlicious~\cite{seizinger2025bokehlicious} obtains the lowest LPIPS (0.1125), indicating the best perceptual quality~\cite{zhang2018unreasonable}. Notably, Bokehlicious~\cite{seizinger2025bokehlicious} also performs particularly well in the most challenging \fnum{2.0} setting, achieving the highest PSNR (24.546) and lowest LPIPS~(0.2043), demonstrating strong robustness under severe defocus blur. For moderate and mild blur (\fnum{4.0} and \fnum{8.0}), other high-capacity transformer- and SSM-based models such as FFTFormer~\cite{kong2023efficient}, EAMamba~\cite{lin2025eamamba}, and LAKDNet~\cite{ruan2023revisiting} also consistently rank among the top performers.

An important observation is the shift in performance trends compared to earlier benchmarks.
With the substantially increased scale, diversity, and quality of RealDefocus~\cite{seizinger2025bokehlicious}, large and expressive architectures, such as Bokehlicious~\cite{seizinger2025bokehlicious}, SSMs (EAMamba~\cite{lin2025eamamba}), and transformer methods (FFTFormer~\cite{kong2023efficient}), can be trained effectively and outperform more traditional designs. This suggests that data scale and coverage play a decisive role in unlocking the capacity of modern high-parameter networks for real-world SIDD.

From an efficiency perspective, lightweight models such as GKMNet~\cite{quan2021gaussian} remain competitive in terms of parameter count (1.41 M) and MACs (20.05 G), but exhibit a clear performance gap compared to the best-performing large models. Conversely, high-capacity methods (e.g., FFTFormer~\cite{kong2023efficient}, Restormer~\cite{zamir2022restormer}, EVSSM~\cite{kong2025efficient}) achieve superior restoration quality at the cost of substantially higher computational demand and inference time. Bokehlicious~\cite{seizinger2025bokehlicious} offers a favorable trade-off, combining competitive complexity (13.96 M parameters, 60.55 GMACs) with state-of-the-art quality.
%% perceptual

Qualitatively, 
as shown in \cref{fig:RealDefocusQual},
the superiority of high-capacity models trained on RealDefocus is reflected in sharper edge reconstruction, improved recovery of fine textures, and reduced ringing or over-smoothing artifacts, particularly in regions with strong depth discontinuities and complex bokeh patterns.

\subsection{Cross-dataset validation on RealDOF}

Table~\ref{tab:RealDOF} reports quantitative results on \textit{RealDOF}~\cite{lee2021iterative} when training on DPDD$_S$~\cite{abuolaim2020defocus} and on RealDefocus~\cite{seizinger2025bokehlicious}. Across all evaluated architectures, training on RealDefocus consistently improves cross-dataset generalization. We observe systematic gains in PSNR and SSIM, accompanied by reduced LPIPS, indicating improved perceptual fidelity.

The improvements are often substantial. For instance, Bokehlicious~\cite{seizinger2025bokehlicious} achieves a +1.402 gain in PSNR and reduces LPIPS by 0.044 when trained on RealDefocus instead of DPDD$_S$. Similarly, FFTFormer~\cite{kong2023efficient} and EAMamba~\cite{lin2025eamamba} benefit from large PSNR gains of +0.772 and +0.971, respectively. Even methods that already generalize relatively well from DPDD$_S$, such as Restormer~\cite{zamir2022restormer} and LAKDNet~\cite{ruan2023revisiting}, consistently improve across all three metrics. 
The qualitative comparison in \cref{fig:RealDOFQual} similarly shows a visible improvement in the fidelity of the restored images.

Overall, these results highlight the superior cross-dataset transferability enabled by RealDefocus~\cite{seizinger2025bokehlicious}. The consistent performance gains across diverse architectures suggest that its increased scale, scene diversity, and aperture coverage lead to models that generalize better to unseen real-world defocus distributions, as represented by RealDOF~\cite{lee2021iterative}.

\section{Discussions}

Current real-world datasets, including RealDefocus~\cite{seizinger2025bokehlicious}, remain limited by aperture-based capture protocols, which leave residual defocus in the ground truth, while light-field cameras are limited in spatial resolution~\cite{ng2006digital}.
Due to a domain gap, training with synthetic Defocus Blur leads to unsatisfying results under real world conditions~\cite{BokehMeHybrid}.

For an improved Defocus Deblurring dataset, one could adopt a focus-stacking approach, where photos with different focus distances are fused into a sharp image~\cite{6738262}.
While current approaches are designed for macro-photography with fixed subjects, it is worth investigating whether this can be applied to deep outdoor scenes to sharpen the ground truth. 

\section{Conclusion}

In this work, we introduce a benchmark protocol for \textit{RealDefocus}, a large-scale real-world dataset for \stask~with standardized splits and evaluation protocols. Our results demonstrate that training on RealDefocus improves cross-dataset generalization versus prior datasets, while enabling unified comparison across task-specific and general restoration architectures, underlining the need for standardized evaluation settings.

% -------------------------------------------------------------------------
{\small
\bibliographystyle{IEEEbib}
\bibliography{strings,refs}
}
\end{document}